\documentclass{article}
\usepackage{iclr2027_conference,times}

\usepackage{amsmath,amsfonts,bm}

\def\eqref#1{equation~\ref{#1}}

\def\1{\bm{1}}

\def\vs{{\bm{s}}}

\DeclareMathAlphabet{\mathsfit}{\encodingdefault}{\sfdefault}{m}{sl}
\SetMathAlphabet{\mathsfit}{bold}{\encodingdefault}{\sfdefault}{bx}{n}

\usepackage{float}
\usepackage{hyperref}
\usepackage{url}
\usepackage{booktabs}
\usepackage{multirow}
\usepackage{xcolor}
\usepackage{colortbl}
\definecolor{oursrowblue}{HTML}{EAF3FC}
\usepackage{graphicx}
\usepackage{afterpage}

\newenvironment{compactequation}[1][2pt]{%
  \setlength{\abovedisplayskip}{#1}%
  \setlength{\belowdisplayskip}{#1}%
  \setlength{\abovedisplayshortskip}{0pt}%
  \setlength{\belowdisplayshortskip}{\dimexpr#1/2\relax}%
  \begin{equation}%
}{\end{equation}\ignorespacesafterend}
\makeatletter
\renewcommand{\section}{\@startsection{section}{1}{\z@}
  {-1.5ex plus -.3ex minus -.2ex}{1.0ex plus .2ex minus .1ex}
  {\large\sc\raggedright}}
\renewcommand{\subsection}{\@startsection{subsection}{2}{\z@}
  {-1.3ex plus -.3ex minus -.2ex}{.6ex plus .1ex}
  {\normalsize\sc\raggedright}}
\renewcommand{\subsubsection}{\@startsection{subsubsection}{3}{\z@}
  {-1.1ex plus -.3ex minus -.2ex}{.4ex plus .1ex}
  {\normalsize\sc\raggedright}}
\renewcommand{\paragraph}{\@startsection{paragraph}{4}{\z@}
  {1.0ex plus .2ex minus .1ex}{-1em}{\normalsize\bf}}
\makeatother

\newcommand{\PlaceScheduledFigures}{%
  \begingroup
    \setlength{\intextsep}{0pt}%
    \ifnum\value{page}=4\relax
\begin{figure}[H]
    \centering
    \includegraphics[width=\linewidth,trim=0 6bp 0 0,clip]{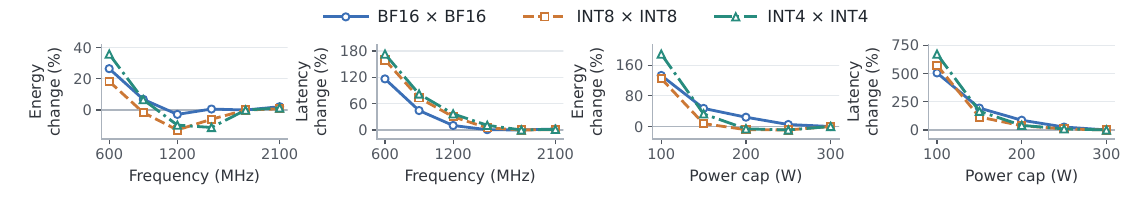}
    \par\vspace{-0.5ex}
    {\small (a) GEMM sensitivity\par}
    \vspace{0ex}
    \includegraphics[width=\linewidth,trim=0 6bp 0 0,clip]{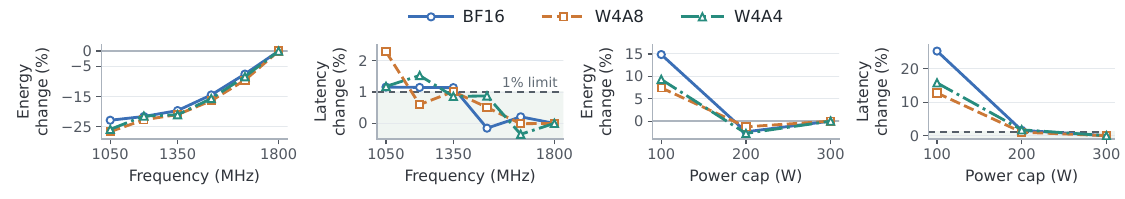}
    \par\vspace{-0.5ex}
    {\small (b) VLA inference sensitivity\par}
    \caption{Energy and latency under frequency and power-cap tuning, relative to 1800 MHz / 300 W. Dashed lines mark the 1\% total-time threshold used in this exploratory sweep.}
    \label{fig:hardware_energy_time_full_grid}
    \vspace{-15pt}
\end{figure}
      \vspace{\textfloatsep}%
    \fi
    \ifnum\value{page}=5\relax
\begin{figure}[H]
    \centering
    \includegraphics[width=\linewidth]{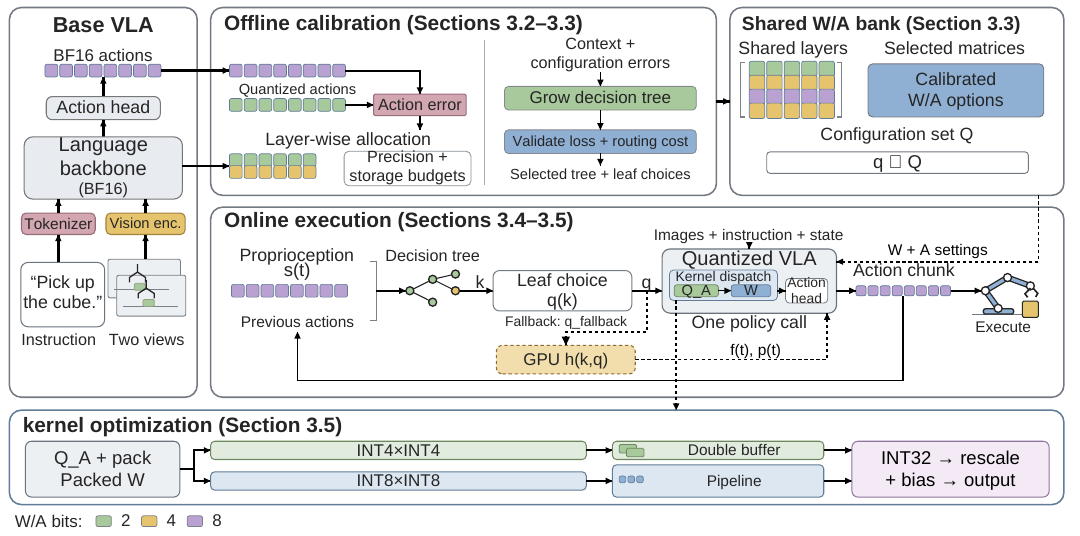}
    \par\vspace{-6pt}
    \caption{Overview of ActTune.}
    \label{fig:phase_conditioned_overview}
\end{figure}
      \vspace{\textfloatsep}%
    \fi
    \ifnum\value{page}=7\relax
\begin{table}[H]
\centering
\caption{Comparison of quantization methods at W4A16, W4A8, and W4A4 on LIBERO for OpenVLA-OFT and $\pi_{0.5}$. Storage includes all persistent tensors. Parentheses show energy with GPU operating-point adaptation (DVFS). Bold marks the best value per model/precision group; energy is compared without DVFS.}
\label{tab:quantization_performance}
\begingroup
\small
\setlength{\tabcolsep}{3pt}
\renewcommand{\arraystretch}{1.08}
\newcommand{\ourscell}{\cellcolor{oursrowblue}}
\resizebox{0.92\textwidth}{!}{%
\begin{tabular}{lllrrrrrrrr}
\toprule
Model & Setting & Method & Spatial & Object & Goal & Long & Mean$\uparrow$ & $\Delta$ & \shortstack{Model storage\\(GB)$\downarrow$} & \shortstack{Energy\\(J/success)$\downarrow$} \\
\midrule
\multirow{12}{*}{\shortstack{OpenVLA\\-OFT}}
 & BF16 & -- & 97.6 & 98.4 & 97.8 & 94.4 & 97.1 & -- & 15.42 & 2830 \\
\cmidrule(lr){2-11}
 & \multirow{3}{*}{W4A16} & QVLA & 96.0 & $\mathbf{98.6}$ & 97.4 & $\mathbf{95.2}$ & 96.8 & $-0.3$ & 4.50 & 3579.1 \\
 & & ActQuant & 96.0 & 98.2 & 97.2 & 95.0 & 96.6 & $-0.5$ & $\mathbf{4.20}$ & 3457.2 \\
 & & \ourscell Ours & \ourscell $\mathbf{97.2}$ & \ourscell 98.0 & \ourscell $\mathbf{97.6}$ & \ourscell 95.0 & \ourscell $\mathbf{97.0}$ & \ourscell $\mathbf{-0.1}$ & \ourscell 4.47 & \ourscell $\mathbf{3369}$ \\
\cmidrule(lr){2-11}
 & \multirow{4}{*}{W4A8} & SmoothQuant & 95.6 & 97.2 & 96.8 & 92.4 & 95.5 & $-1.6$ & 6.32 & 2057.5 \\
 & & GPTQ & 97.4 & 96.8 & $\mathbf{98.2}$ & 92.8 & 96.3 & $-0.8$ & 6.01 & 1944.8 \\
 & & AWQ & $\mathbf{98.4}$ & 96.8 & 97.2 & 93.4 & 96.5 & $-0.6$ & 5.91 & 1986.7 \\
 & & \ourscell Ours & \ourscell 97.4 & \ourscell $\mathbf{98.2}$ & \ourscell 97.6 & \ourscell $\mathbf{94.8}$ & \ourscell $\mathbf{97.0}$ & \ourscell $\mathbf{-0.1}$ & \ourscell $\mathbf{4.47}$ & \ourscell $\mathbf{1145.3}$ (843.1) \\
\cmidrule(lr){2-11}
 & \multirow{4}{*}{W4A4} & SmoothQuant & 77.2 & 70.0 & 77.8 & 68.6 & 73.4 & $-23.7$ & 6.32 & 2763.7 \\
 & & GPTQ & 91.4 & 94.4 & 92.2 & 87.8 & 91.5 & $-5.6$ & 6.01 & 1453.9 \\
 & & AWQ & 94.2 & 90.4 & 72.4 & 16.8 & 68.5 & $-28.6$ & 5.91 & 2788.2 \\
 & & \ourscell Ours & \ourscell $\mathbf{96.8}$ & \ourscell $\mathbf{97.6}$ & \ourscell $\mathbf{96.8}$ & \ourscell $\mathbf{90.8}$ & \ourscell $\mathbf{95.5}$ & \ourscell $\mathbf{-1.6}$ & \ourscell $\mathbf{4.47}$ & \ourscell $\mathbf{1135.4}$ (905.0) \\
\midrule
\multirow{12}{*}{$\pi_{0.5}$}
 & BF16 & -- & 98.4 & 98.4 & 97.6 & 93.6 & 97.0 & -- & 7.47 & 2836 \\
\cmidrule(lr){2-11}
 & \multirow{3}{*}{W4A16} & QVLA & 98.0 & 97.2 & 96.4 & 91.8 & 95.9 & $-1.2$ & 6.3 & 3468.5 \\
 & & ActQuant & 98.4 & $\mathbf{99.4}$ & 96.8 & 91.8 & 96.6 & $-0.4$ & 6.20 & 3364.9 \\
 & & \ourscell Ours & \ourscell $\mathbf{99.4}$ & \ourscell $\mathbf{99.4}$ & \ourscell $\mathbf{98.6}$ & \ourscell $\mathbf{95.0}$ & \ourscell $\mathbf{98.1}$ & \ourscell $\mathbf{+1.1}$ & \ourscell $\mathbf{3.42}$ & \ourscell $\mathbf{3355}$ \\
\cmidrule(lr){2-11}
 & \multirow{4}{*}{W4A8} & SmoothQuant & 98.8 & 99.4 & 98.2 & $\mathbf{97.6}$ & $\mathbf{98.5}$ & $\mathbf{+1.5}$ & 3.67 & 1548.8 \\
 & & GPTQ & 99.0 & $\mathbf{99.6}$ & $\mathbf{98.8}$ & 94.6 & 98.0 & $+1.0$ & 3.57 & 1722.6 \\
 & & AWQ & 98.0 & 99.2 & 98.2 & 95.8 & 97.8 & $+0.8$ & 3.53 & 1662.9 \\
 & & \ourscell Ours & \ourscell $\mathbf{99.4}$ & \ourscell 98.6 & \ourscell 98.0 & \ourscell 96.2 & \ourscell 98.1 & \ourscell $+1.1$ & \ourscell $\mathbf{3.42}$ & \ourscell $\mathbf{825.1}$ (657.2) \\
\cmidrule(lr){2-11}
 & \multirow{4}{*}{W4A4} & SmoothQuant & 93.6 & 98.8 & 71.8 & 58.0 & 80.6 & $-16.5$ & 3.67 & 1902.5 \\
 & & GPTQ & 96.6 & $\mathbf{99.6}$ & 91.4 & 85.6 & 93.3 & $-3.7$ & 3.57 & $\mathbf{1341.7}$ \\
 & & AWQ & 72.4 & 94.2 & 49.4 & 27.0 & 60.8 & $-36.3$ & 3.53 & 2965.3 \\
 & & \ourscell Ours & \ourscell $\mathbf{99.4}$ & \ourscell $\mathbf{99.6}$ & \ourscell $\mathbf{98.8}$ & \ourscell $\mathbf{95.8}$ & \ourscell $\mathbf{98.4}$ & \ourscell $\mathbf{+1.4}$ & \ourscell $\mathbf{3.42}$ & \ourscell 1527.2 (1221.8) \\
\bottomrule
\end{tabular}%
} 
\endgroup
\vspace{2pt}
\end{table}
      \vspace{4pt}
    \fi
    \ifnum\value{page}=8\relax
      \begin{figure}[H]
\centering
\includegraphics[width=\linewidth]{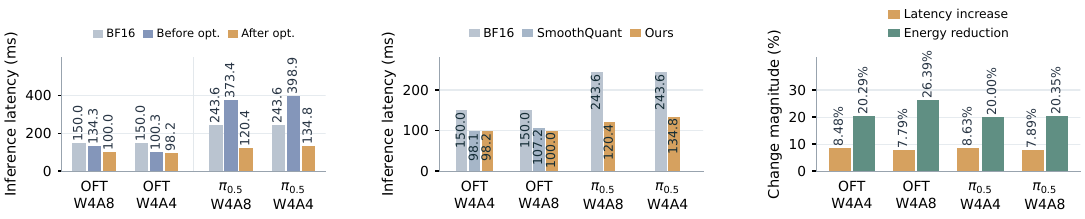}
\caption{Inference latency before/after optimization (left) and across methods (middle), and DVFS latency overhead and energy-per-success savings relative to 1800\,MHz / 300\,W (right).}
\label{fig:inference_runtime_comparison}
\end{figure}
      \vspace{\textfloatsep}%
    \fi
    \ifnum\value{page}=9\relax
      \begin{figure}[H]
\centering
\begin{minipage}[b]{0.27\linewidth}
\centering
\includegraphics[width=\linewidth]{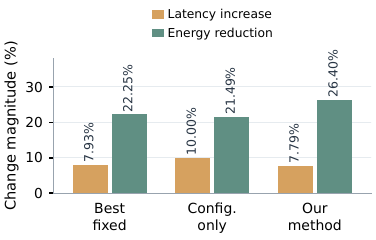}
\par\smallskip
{\small (a) GPU settings}
\end{minipage}\hfill
\begin{minipage}[b]{0.39\linewidth}
\centering
\includegraphics[width=\linewidth]{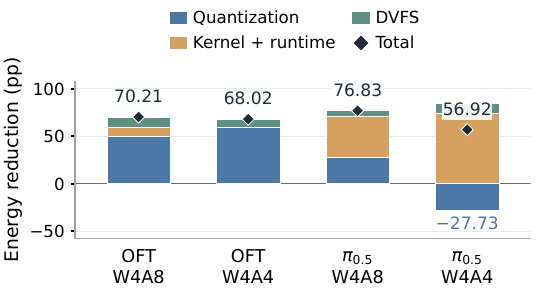}
\par\smallskip
{\small (b) Energy contributions}
\end{minipage}\hfill
\begin{minipage}[b]{0.32\linewidth}
\centering
\includegraphics[width=\linewidth]{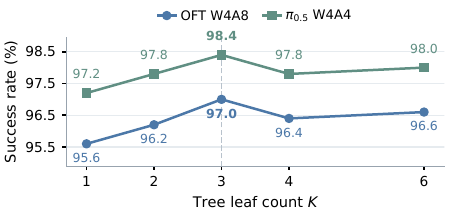}
\par\smallskip
{\small (c) Closed-loop success}
\end{minipage}
\caption{ActTune ablations: (a) GPU operating-point policies; (b) energy-saving contributions in percentage points of BF16 energy per success; (c) closed-loop success versus leaf count $K$.} 
\label{fig:hardware_ablation}
\vspace{-10pt}
\end{figure}
      \vspace{\textfloatsep}%
    \fi
  \endgroup
  \ifnum\value{page}<9\relax
    \afterpage{\PlaceScheduledFigures}%
  \fi
}

\title{ActTune: Action-Aware Precision and GPU Operating-Point Adaptation for Energy-Efficient Vision-Language-Action Inference}
\author{%
  \textbf{Zou Qingyun\textsuperscript{1}\quad Bin Gao\textsuperscript{1}\quad Wenju Zhao\textsuperscript{2}}\\
  \textbf{Weng-Fai Wong\textsuperscript{1}\quad Bingsheng He\textsuperscript{1}\quad Tulika Mitra\textsuperscript{1}}\\[0.5em]
  \normalfont\textsuperscript{1}National University of Singapore\\
  \normalfont\textsuperscript{2}Huazhong University of Science and Technology
}
\iclrfinalcopy
\hypersetup{
  hidelinks,
  pdftitle={ActTune: Action-Aware Precision and GPU Operating-Point Adaptation for Energy-Efficient Vision-Language-Action Inference},
  pdfauthor={Zou Qingyun, Bin Gao, Wenju Zhao, Weng-Fai Wong, Bingsheng He, Tulika Mitra}
}

\begin{document}
\maketitle
\lhead{Under review as a conference paper at ICLR 2027}
\afterpage{\PlaceScheduledFigures}

\begin{abstract}
Vision-language-action (VLA) policies repeatedly invoke inference to control robots, making graphics processing unit (GPU) energy a recurring cost of task execution. Reducing energy per inference call, however, may not reduce energy per successful task if numerical errors increase failures or slower inference prolongs execution. We therefore target GPU energy per successful task while preserving task success and keeping the inference-latency increase within 10\%. Our approach builds on two observations: quantization sensitivity varies across action classes, model layers, and weights versus activations; and numerical precision changes the workload, shifting favorable GPU operating points. We introduce ActTune, an action-aware framework that connects layer-wise precision allocation with workload-dependent GPU operating-point selection over requested frequency--power-cap pairs. A lightweight decision tree learns its splits and leaf precision configurations directly from configuration action errors, then selects precision before each policy call. The controller forecasts the next workload and applies the selected GPU operating point asynchronously using a lookup table calibrated under a latency budget. A shared resident quantized weight bank enables configuration switching without weight reconstruction or additional policy evaluations. On LIBERO, a benchmark for lifelong robot learning, ActTune improves mean task success by up to 2.3\% relative to state of the art. Relative to the original BF16 implementations, it delivers up to $2.02\times$ faster inference and, with GPU operating-point adaptation, reduces energy per successful task by up to 76.8\%.
\end{abstract}

\section{Introduction}
\label{sec:introduction}


VLA models translate visual observations and language instructions into robot actions~\citep{kim2024openvla}. For battery-powered robots, repeated onboard inference consumes limited battery energy and can shorten operation between charges. Reducing inference energy is therefore important for extending robot endurance. Dynamic voltage and frequency scaling (DVFS) offers a direct way to pursue this goal by lowering power consumption~\citep{tang2019dvfs}. However, lower frequencies prolong inference, which can offset energy savings and delay action updates. The need for timely responses therefore limits how aggressively frequency can be reduced.

Numerical precision and GPU operating-point control offer complementary ways to reduce this cost. Quantization lowers weight and activation precision, changing computation and memory traffic, but the resulting numerical errors can also perturb predicted actions. We define a GPU operating point as a requested frequency--power-cap pair. Adjusting these settings trades execution speed against power consumption. Lower frequencies may reduce power, yet longer inference can offset these savings and delay the robot's response to new observations. Crucially, these two controls are coupled: precision determines the workload on which GPU operating-point control operates. An energy-efficient operating point for one precision configuration may therefore be less effective for another. Exploiting both controls requires connecting the precision needs of robot actions to the hardware characteristics of the resulting inference workload.

Existing work provides useful foundations for this connection. VLA quantization methods use action or task sensitivity to allocate precision~\citep{xu2026qvla,akbari2026actquant,ranjan2026mixqvla}, while temporal-dynamic-aware quantization for VLA models (DyQ-VLA) and motion-aware vector quantization (MotionVQ) adapt precision during execution using motion signals~\citep{zheng2026dyqvla,song2026vqvla}. However, compression objectives such as model size and bit-operation budgets do not directly capture GPU energy during task execution~\citep{ranjan2026mixqvla}. Other work studies energy efficiency through quantized robotic policies or device--edge execution placement~\citep{park2025sqil,zhou2026ecovla}, and PiQi, a framework for partially quantized inference, combines partial quantization, processor placement, and DVFS for general deep neural network (DNN) inference~\citep{aghapour2024piqi}. Building on these directions, we investigate how action-aware precision allocation and workload-dependent GPU operating-point selection can work together to reduce energy per successful robot task.

\begin{figure}[t]
\centering
\includegraphics[width=\linewidth]{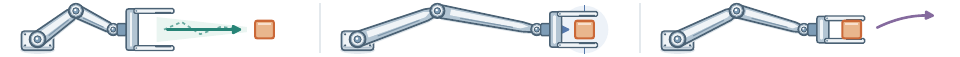}
\par\smallskip
\noindent\makebox[0.333333\linewidth][c]{\small (a) Coarse approach}%
\makebox[0.333333\linewidth][c]{\small (b) Final alignment}%
\makebox[0.333333\linewidth][c]{\small (c) Object transport}
\caption{ActTune intuition: approach tolerates errors, alignment requires accuracy, and transport precision depends on path and grip stability.} 
\label{fig:action_precision_intuition}
\vspace{-10pt}
\end{figure}

Our approach is motivated by two observations. First, quantization sensitivity depends on both the action being performed and where numerical error is introduced in the model. For example, coarse approach motion may tolerate small deviations, whereas final alignment may require greater accuracy (Figure~\ref{fig:action_precision_intuition}). Using a fixed set of action classes learned from training actions, short quantized interventions reveal sensitivity differences across action classes, Transformer blocks, and weight versus activation precision (Figure~\ref{fig:motivation_action_combined}(b--c)). Second, hardware measurements show that different numerical precisions can favor different GPU operating points (Figure~\ref{fig:hardware_energy_time_full_grid}). Together, these observations suggest a two-stage adaptation strategy: use action sensitivity to guide precision allocation, then use the resulting workload to guide GPU operating-point selection.

We introduce \textbf{ActTune}, an action-aware framework that implements this strategy. Before each action chunk is generated, a lightweight decision tree uses current proprioception and the preceding predicted action chunk to select a layer-wise weight and activation precision configuration. Offline calibration learns the tree splits and leaf configurations directly from action error against the BF16 reference policy under precision and storage budgets. A shared resident quantized weight bank supports configuration switching without reconstructing weights or storing a complete model copy for every leaf. This design enables action-aware precision selection with only one policy inference call per action chunk.

ActTune profiles frequency and power-cap settings on replayed training workloads with the precision selector fixed. It selects the candidate table with the lowest estimated energy while limiting the mean inference-latency increase to 10\% relative to the same precision selector at the fixed reference GPU operating point, subject to action equality and memory checks. After each call, the controller forecasts the next workload and applies the selected GPU operating point asynchronously. Precision selection uses the current observation. Because matching actions on replayed inputs does not guarantee unchanged closed-loop behavior, the final assessment must consider complete episodes. We therefore evaluate closed-loop task success and energy per successful task, and check the 10\% budget using the measured inference latency.
Our contributions are:
\begin{list}{\textbullet}{%
    \setlength{\leftmargin}{0pt}%
    \setlength{\itemindent}{0pt}%
    \setlength{\labelwidth}{1em}%
    \setlength{\labelsep}{0.5em}%
}
    \item \textbf{Empirical characterization.} We firstly characterize how quantization sensitivity varies across action classes, Transformer blocks, and weight versus activation precision, and show that favorable GPU operating points depend on numerical precision.

    \item \textbf{Action-aware precision and GPU operating-point adaptation.} We propose ActTune, which connects action-conditioned layer-wise precision allocation with workload-dependent GPU operating-point selection. A lightweight decision tree and a shared quantized weight bank enable online adaptation with one policy call per action chunk.

    \item \textbf{Quantitative evaluation.} In the evaluated LIBERO comparisons, ActTune improves mean task success by up to 2.3\% relative to QVLA, an action-centric VLA quantization framework, and action-guided quantization (ActQuant). It also reduces static model storage by up to 29.3\% relative to the compared general-purpose quantization baselines at W4A8/W4A4 (Table~\ref{tab:quantization_performance}).
    Compared with the original BF16 implementations, it achieves up to $2.02\times$ inference speedup and up to 76.8\% lower energy per successful task with GPU operating-point adaptation.
\end{list}

\vspace{-8pt}
\section{Motivation}
\label{sec:motivation}

We examine two empirical questions that motivate ActTune:
(1) whether quantization sensitivity depends on the action and on where precision is reduced in the model, and
(2) whether different precision configurations favor different GPU operating points.
The first determines how precision should be allocated, while the second determines whether hardware settings should adapt to the resulting workload.

\begin{figure}[t]
\centering
\includegraphics[width=\linewidth,trim=0 22bp 0 0,clip]{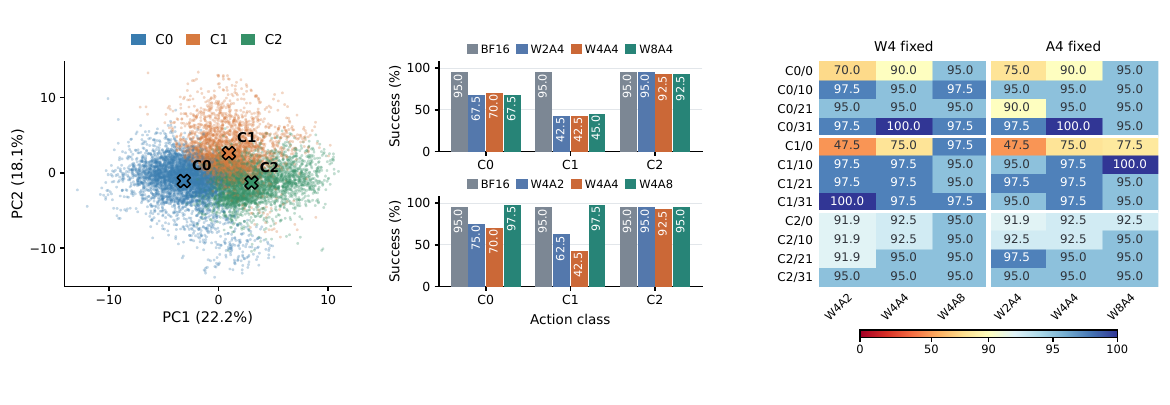}
\par\vspace{-0.5ex}
\noindent\hbox to \linewidth{%
\hspace{0.02\linewidth}%
\makebox[0.32\linewidth][c]{\small (a) Action clusters} 
\makebox[0.295\linewidth][c]{\small (b) Action sensitivity}%
\hspace{0.055\linewidth}%
\makebox[0.31\linewidth][c]{\small (c) Layer sensitivity}%
}
\vspace{-5pt}
\caption{Action grouping and precision sensitivity. PC1 and PC2 denote the first and second principal components. W$n$A$m$ denotes $n$-bit weights and $m$-bit activations. Crosses mark cluster centroids; success rates average all four suites, with all action and layer precision batches complete.}
\label{fig:motivation_action_combined}
\vspace{-15pt}
\end{figure}


\paragraph{Action- and layer-dependent quantization sensitivity.}
We first group training action blocks into three fixed classes, C0--C2, using principal component analysis (PCA) followed by K-means (Figure~\ref{fig:motivation_action_combined}(a)).
These classes describe patterns in the action data rather than prescribed physical task stages.
To measure quantization sensitivity, we insert an eight-action quantized intervention into matched BF16 trajectories and then return to BF16 control.
We measure final task success rate using equal-weight averages over Goal, Object, Spatial, and Long.
The matched BF16 reference achieves a mean task success rate of 95.0\%. Figure~\ref{fig:motivation_action_combined}(b) shows that the same precision configuration can have substantially different effects across action classes.
At fixed A4, C1 achieves only 42.5--45.0\% task success across the tested weight precisions, whereas C2 maintains 92.5--95.0\%.
For C1, changing activation precision has a larger effect than changing weight precision, and increasing weight precision alone does not resolve the poor A4 outcomes. Sensitivity is also nonuniform across Transformer layers (Figure~\ref{fig:motivation_action_combined}(c)).
For C1, applying W4A4 to layer 0 yields 75.0\% task success, compared with 97.5\% when applying the same precision to the other tested layers.
Increasing layer-0 activation precision from A4 to A8 raises task success to 97.5\%.
These results show that precision requirements depend jointly on the action class, model location, and quantized component, motivating action-conditioned, layer-wise allocation of weight and activation precision.

\paragraph{Precision-dependent GPU energy--latency tradeoffs.}
We next test whether numerical precision changes the preferred GPU operating point.
On an NVIDIA RTX A6000, we sweep requested GPU frequency at a fixed 300\,W power cap and sweep power cap at a requested frequency of 1800\,MHz (Figure~\ref{fig:hardware_energy_time_full_grid}).
Measurements for each precision are normalized to its own 1800\,MHz / 300\,W reference, isolating the effect of hardware tuning from the energy differences introduced by quantization. At the kernel level, we benchmark resident INT4$\times$INT4, INT8$\times$INT8, and BF16$\times$BF16 GEMM workloads.
The measurements include the output epilogue but exclude input quantization, packing, and transfers. INT4 and INT8 reach their lowest measured GPU energy at different frequencies (Figure~\ref{fig:hardware_energy_time_full_grid}(a)), showing that the energy--latency tradeoff depends on numerical precision. An exploratory policy-level sweep shows the same tendency during VLA execution (Figure~\ref{fig:hardware_energy_time_full_grid}(b)).
In this exploratory sweep, among measured operating points that increase total benchmark time by at most 1\%, the lowest-energy points reduce GPU energy by 20.4\%, 22.7\%, and 21.0\% for BF16, W4A8, and W4A4, respectively, relative to their corresponding reference operating points.
These points use different frequency--power-cap combinations, motivating configuration-specific hardware profiling rather than a common operating point for all precision settings.

The clusters above serve as diagnostics; ActTune learns decision regions from each configuration's action error (Section~\ref{sec:action_class_recognition}).
The hardware sweep uses four matched episodes and one measurement per setting, so it provides design evidence rather than a precise estimate of task-level savings.
These findings motivate action-aware precision allocation and workload-dependent GPU control.

\section{The ActTune Design}
\label{sec:method}

\subsection{Overview}
\label{sec:acttune-overview}

ActTune adapts precision and GPU operating points to reduce energy per successful task while preserving task success and limiting the inference-latency increase from operating-point adaptation to 10\%. Following Section~\ref{sec:motivation}, precision selection accounts for action-dependent quantization sensitivity, while GPU frequency and power cap depend on the resulting workload. The controller selects precision before each policy call and prepares GPU operating points asynchronously between calls by forecasting the next workload. At request \(t\), the policy maps observation \(o_t\), comprising images, a language instruction, and proprioception \(s_t\in\mathbb{R}^{d_s}\), to an action chunk \(A_t\in\mathbb{R}^{H\times7}\). Each action contains six motion coordinates and a gripper command. A GPU operating point \(u_t=(f_t,p_t)\) pairs the requested frequency \(f_t\) with power cap \(p_t\). The selected precision configuration \(q_t\) and operating point \(u_t\) remain fixed throughout inference, including iterative action generation. The predicted chunk provides action-history features for the next request. Figure~\ref{fig:phase_conditioned_overview} shows the workflow. Offline calibration learns candidate decision trees from configuration errors and selects their complexity using validation loss and routing cost (Sections~\ref{sec:action_class_recognition}--\ref{sec:shared_bank}). Each leaf stores a precision configuration from the shared weight bank. With the tree fixed, GPU profiling selects operating points under a latency budget (Section~\ref{sec:frequency_selection}). Online execution uses one policy call per request (Section~\ref{sec:online_energy}).

\subsection{Learning decision regions for precision selection}
\label{sec:action_class_recognition}

ActTune learns when to use each precision configuration from calibration errors. A binary tree partitions action context into decision regions, with one precision configuration stored at each leaf. Calibration learns the split conditions and configuration choices, while validation characterizes the tradeoff between the number of regions and routing cost. The method does not prescribe a fixed number of leaves. The tree forms \(x_t=[\phi(A_{t-1},m_{t-1});s_t]\) from current proprioception and the preceding predicted chunk, then selects leaf \(k_t=g(x_t)\in\{1,\ldots,K\}\) and configuration \(q_t=q_{k_t}\). Here \(\phi\) extracts motion and gripper features, \(m_{t-1}\) is the preceding gripper margin, \(g(x_t)\) is the reached leaf, and \(q_k\) is its configuration (Appendix~\ref{app:online_recognizer}). The selected number of leaves is \(K\). These inputs are available before inference, so selection requires no additional policy call. Regions are learned from configuration losses, not predefined semantic action stages; the tree depth bounds the number of online comparisons.

\subsection{Action-conditioned precision with shared resident weights}
\label{sec:shared_bank}
\label{sec:class_mapping}

ActTune allocates backbone precision under average weight and activation budgets, weighted by parameter count and input element count, respectively. Individual matrices may use different precisions within these budgets. A separate resident-storage bound counts all retained weight versions and quantization metadata, including inactive alternatives (Appendix~\ref{app:precision_storage_budgets}).

\paragraph{Shared configurations.} Offline calibration selects switchable matrices and their precision options under the precision and resident-storage budgets. These options form a configuration set \(\mathcal Q\); all other targeted matrices share one resident version. Each tree leaf selects a configuration, and multiple leaves may share it, so leaf count and configuration count need not match. Configurations specify resident weight versions and runtime activation precision. Calibration also determines a fallback configuration \(q_{\mathrm{fallback}}\in\mathcal Q\).

\paragraph{Sensitivity-guided bank construction.} We calibrate layer alternatives against the BF16 checkpoint using activation-guided channel scaling based on activation-aware weight quantization (AWQ)~\citep{lin2023awq} and output-row clipping to minimize local reconstruction error. Weighting these errors by the action Jacobian prioritizes matrices whose quantization affects action predictions more strongly. Matrices with low penalties receive fewer weight bits, freeing budget for upgrades with greater error reduction per added bit. We lower shared weight precision as needed to meet the resident-storage bound, prioritizing small error penalties per byte saved. We calibrate candidate weight/activation pairs and select a feasible bank using training action error (Appendix~\ref{app:layer_precision_selection}). For each model--budget pair, one tree shares its splits and configuration indices across tasks; each index resolves to the bank associated with the task's checkpoint.

\paragraph{Learning the partition and configuration choices.} For each calibration observation \(i\), we evaluate every configuration in \(\mathcal Q\) against the same BF16 reference. With postprocessed action chunks \(A_i^q\) and \(A_i^{\mathrm{ref}}\), the loss is:
\begin{compactequation}[0pt]
\ell_i(q)=\frac{1}{6H}\sum_{h=1}^{H}\sum_{d=1}^{6}
\left|A_{i,h,d}^{q}-A_{i,h,d}^{\mathrm{ref}}\right|
+\frac{\lambda_g}{H}\sum_{h=1}^{H}
\mathbf{1}\!\left[A_{i,h,7}^{q}\neq A_{i,h,7}^{\mathrm{ref}}\right].
\label{eq:action_calibration_loss}
\end{compactequation}
Here \(\lambda_g\) weights gripper disagreement. For \(N\) fitting contexts and each candidate leaf count \(K\), we optimize action agreement:
\begin{compactequation}[0pt]
\min_{g,\,\{q_k\in\mathcal Q\}}
\sum_{i=1}^{N}w_i\,\ell_i\!\left(q_{g(x_i)}\right)
\quad\text{subject to}\quad |\operatorname{Leaves}(g)|=K.
\label{eq:tree_calibration}
\end{compactequation}
Weights \(w_i\) sum to one and assign equal mass to suites, then trajectories within each suite, then valid contexts within each trajectory. Each leaf chooses the configuration with the lowest weighted fitting loss. We grow a tree by loss-reducing splits, then use bottom-up dynamic programming to obtain the lowest-fitting-loss subtree for each feasible leaf count (Appendix~\ref{app:class_mapping_details}).

\paragraph{Selecting complexity and freezing.} We select tree size by comparing validation action loss and measured CPU routing cost across candidate leaf counts, including a single-leaf global selector. Entire calibration trajectories are held out before fitting the bank and tree; validation leaves the fitted splits and leaf configurations unchanged. For a common leaf count across model--budget pairs, we compare their validation curves jointly. Section~\ref{sec:ablation_study} reports the resulting experimental choice. We freeze the chosen tree before subsequent GPU profiling. Calibration uses the preceding fallback-policy chunk as history; deployment uses the selected configuration's output. The closed-loop evaluations in Section~\ref{sec:experiments} assess this history shift and task success using the frozen tree.

\subsection{Selecting GPU frequency and power cap}
\label{sec:frequency_selection}

Precision changes computation and memory traffic, shifting favorable GPU operating points (Section~\ref{sec:motivation}). We therefore profile frequency and power cap jointly, with the weight bank and precision selector fixed. The experiments label this joint operating-point control policy ``DVFS''; its controls are requested GPU frequency and power cap.

\paragraph{Offline profiling.} For each supported region and configuration pair \((k,q)\), we measure energy and latency on training observations at operating points \(u=(f,p)\in\mathcal U\), relative to \(u_{\mathrm{ref}}=(f_{\mathrm{ref}},p_{\mathrm{ref}})\). Comparisons use the same memory frequency, observations, selector states, and random states for stochastic policies. Closed-loop measurements also include switching overhead. We retain operating points that preserve reference actions and pass memory checks. Uniform settings and a sweep over latency penalties define candidate lookup tables, each mapping a region--configuration pair to a GPU operating point. Using replay measurements, we select the table \(h^*\) with the lowest estimated GPU energy whose mean inference latency increases by at most 10\% relative to the same precision selector at \(u_{\mathrm{ref}}\). Unsupported pairs use calibrated fallbacks. We freeze the table, verify actions and memory on disjoint trajectories, then evaluate closed-loop task success, energy per success, and inference latency on the complete benchmark. The 10\% budget applies to mean complete prediction-call latency, measured against the same frozen precision selector at the fixed reference operating point \(u_{\mathrm{ref}}\). Figure~\ref{fig:inference_runtime_comparison} reports this comparison for all four model--precision settings. Energy per success is total episode GPU energy, including failures and gaps between calls, divided by successes.

\paragraph{Online use.} After each inference, the controller predicts the next decision region and precision configuration. It looks up the corresponding GPU frequency and power cap in the frozen table and applies them asynchronously between calls. These settings remain fixed during the next inference.

\subsection{Online execution and integer kernels}
\label{sec:online_energy}
\label{sec:a6000_kernels}

Each episode starts with empty action history and the fallback operating point. Before inference, the controller waits for any pending GPU update, traverses the frozen tree with \(x_t\), and reads \(q_t=q_{g(x_t)}\). Missing history or invalid input features use \(q_{\mathrm{fallback}}\). The policy produces \(A_t=\pi_{q_t}(o_t)\) at the active operating point. After inference, the controller uses new action features and bounded proprioception extrapolation to forecast the next region and precision configuration. It applies the operating point from \(h^*\) asynchronously between calls. Forecast errors leave the prepared operating point active; precision uses the current observation. Episode measurements include switching and remaining waits. Appendix~\ref{app:execution_scope} gives control and runtime details.

\noindent\textbf{Executing the selected precision.} ActTune executes low-bit linear layers without reconstructing dense weights. It uses INT4\(\times\)INT4 or INT8\(\times\)INT8 Tensor Core kernels with INT32 accumulation~\citep{nvidia2022ptx}. Quantization bit width determines represented values and resident weight storage; the kernel determines arithmetic operand width. Extending a code to a wider operand preserves its value and quantization precision.

Layers whose weights and activations both use at most four bits run with INT4 arithmetic, extending two-bit codes to signed INT4 as needed. Layers with eight-bit weights or activations use INT8 arithmetic, extending narrower codes inside the kernel. W2/W4 weights for A8 layers remain packed in device memory and are sign-extended into INT8 tiles during GEMM. Together, these paths cover all nine local weight/activation pairs without changing the calibrated codes. The INT4 kernel, implemented in NVIDIA's Compute Unified Device Architecture (CUDA) programming platform, uses packed operands, shape-dependent tiles, asynchronous copies, double-buffered shared memory, and fragment reuse. The INT8 Triton kernel uses coalesced weight layouts, tile-local unpacking, and software pipelining. Fusing activation quantization and GEMM epilogues for rescaling, bias addition, and output conversion reduces kernel launches and intermediate memory traffic. Policy execution includes these operations; the resident-GEMM measurements in Section~\ref{sec:motivation} exclude input quantization, packing, and transfers.
By shortening execution time, these kernel and runtime optimizations also help reduce energy consumption, as shown by the energy-saving contributions in Figure~\ref{fig:hardware_ablation}b.

\section{Experiments}
\label{sec:experiments}

\subsection{Experimental setup}
\label{sec:experimental_setup}

We evaluate quantized OpenVLA-OFT~\citep{kim2025oft} and $\pi_{0.5}$~\citep{physicalintelligence2025pi05} on the LIBERO Spatial, Object, Goal, and Long suites~\citep{liu2023libero} (Table~\ref{tab:quantization_performance}).
Our hardware measurements use NVIDIA RTX A6000 GPUs.
We compare ActTune with QVLA~\citep{xu2026qvla} and ActQuant~\citep{akbari2026actquant} at W4A16, and with SmoothQuant~\citep{xiao2022smoothquant}, GPTQ~\citep{frantar2022gptq}, and AWQ~\citep{lin2023awq} at W4A8 and W4A4.
We quantize eligible linear layers in both the vision encoder and language Transformer; other components retain their existing precision.
We report task success, inference speedup, compressed-model storage, and GPU-board energy per successful episode. Appendix~\ref{app:evaluation_settings} specifies the evaluation protocol and selector settings.
Appendices~\ref{app:experimental_details}--\ref{app:supplementary_results} give hardware and controller settings, calibration details, and supplementary results.

We use the average-precision budgets in Section~\ref{sec:shared_bank}. Two switchable matrices, each with two precision options, give \(\mathcal Q=\{q_{00},q_{01},q_{10},q_{11}\}\), with \(q_{\mathrm{fallback}}=q_{11}\). Appendix~\ref{app:layer_precision_selection} gives the selected matrices, precision options, and calibration procedure. For each model--budget pair, one tree is shared across the four LIBERO suites. OFT uses the matching suite checkpoint and its calibrated bank; $\pi_{0.5}$ shares both checkpoint and bank across suites.

\subsection{Results}
\label{sec:quantization_performance}

Table~\ref{tab:quantization_performance} compares task success, static model storage, and energy per successful episode at W4A16, W4A8, and W4A4 on the four LIBERO suites.
Figure~\ref{fig:inference_runtime_comparison} shows inference latency before and after optimization (left), latency comparisons (middle), and latency overhead and energy savings from GPU operating-point adaptation (right).

\noindent\textbf{General-purpose quantization comparison.}
General-purpose methods show larger success losses at W4A4 than at W4A8 (Table~\ref{tab:quantization_performance}).
At W4A4, ActTune reaches 95.5\% on OpenVLA-OFT and 98.4\% on $\pi_{0.5}$, exceeding the strongest compared baseline by 4.0 and 5.1 percentage points, respectively.
This gap is consistent with preserving precision in action-sensitive computations.
ActTune also has the smallest reported storage in these groups (4.47\,GB and 3.42\,GB, including all resident weight versions) and the lowest W4A8 energy per successful episode on both models at the fixed reference GPU operating point.




\noindent\textbf{VLA quantization comparison.}
At W4A16, ActTune reaches 97.0\% mean success on OpenVLA-OFT and 98.1\% on $\pi_{0.5}$, exceeding QVLA~\citep{xu2026qvla} and ActQuant~\citep{akbari2026actquant} on both models (Table~\ref{tab:quantization_performance}).
On $\pi_{0.5}$, it also reduces storage to 3.42\,GB, 44.8--45.7\% below these baselines, combining higher success with a smaller model.
On OpenVLA-OFT, storage is similar: 4.47\,GB versus 4.50\,GB for QVLA and 4.20\,GB for ActQuant.
Energy per successful episode is 2.6--5.9\% lower on OpenVLA-OFT and 0.3--3.3\% lower on $\pi_{0.5}$ than these baselines, but remains above BF16 on both models.

\noindent\textbf{Inference speedup.}
Figure~\ref{fig:inference_runtime_comparison} (left) shows that runtime optimization reduces inference latency across all four configurations, with the largest gains on $\pi_{0.5}$.
Its W4A8 and W4A4 implementations change from slower than BF16 to $2.02\times$ and $1.81\times$ faster, while OpenVLA-OFT reaches $1.50\times$ and $1.53\times$, respectively.
Figure~\ref{fig:inference_runtime_comparison} (middle) compares ActTune inference latency with baseline reference values.

\noindent\textbf{Energy analysis.}
At the fixed reference GPU operating point, ActTune's W4A8/W4A4 configurations reduce energy per successful episode relative to BF16 by 59.5\%/59.9\% on OpenVLA-OFT and 70.9\%/46.1\% on $\pi_{0.5}$ (Table~\ref{tab:quantization_performance}).
With GPU operating-point adaptation enabled, the parenthesized energy entries in Table~\ref{tab:quantization_performance} correspond to reductions of 70.2\%/68.0\% and 76.8\%/56.9\%, respectively.
Figure~\ref{fig:inference_runtime_comparison} (right) shows that GPU operating-point adaptation increases mean inference latency by 7.79--8.63\% relative to the same quantized policies at the fixed 1800\,MHz / 300\,W reference operating point. All four model--precision settings remain within the 10\% inference-latency budget.

\subsection{Ablation Study}
\label{sec:ablation_study}


\paragraph{GPU operating-point selection.}
We hold the precision selector fixed and compare four GPU operating-point policies on 400 episodes, sampling 100 from each LIBERO suite (Figure~\ref{fig:hardware_ablation}a).
Appendix~\ref{app:hardware_ablation_cohort} describes the sampling procedure.
Our method saves 26.40\% energy relative to the reference, exceeding the best fixed setting and configuration-only control by 4.15 and 4.91 percentage points, respectively.
Its inference latency increase is 7.79\%, compared with 7.93\% and 10.00\% for these alternatives.
These results support using decision-region information in hardware selection, while showing that configuration-only adaptation does not necessarily outperform a well-chosen fixed setting.

\paragraph{Energy-saving contributions.}
Figure~\ref{fig:hardware_ablation}b reports energy-saving contributions measured through sequential ablations of quantization, kernel and runtime optimization, and GPU operating-point adaptation (DVFS). Each contribution is the measured energy-per-success decrease from the preceding variant, normalized by BF16 energy per success; Appendix~\ref{app:energy_contributions} details the ablation and normalization.
Quantization contributes 49.82 and 59.27 percentage points on OFT W4A8 and W4A4, while kernel/runtime optimization contributes 42.82 and 73.88 points on $\pi_{0.5}$ W4A8 and W4A4.
Lower power does not necessarily mean lower energy: energy integrates power over execution time, so a sufficiently long execution can outweigh a reduction in average power. For $\pi_{0.5}$ W4A4, quantization alone increases energy per success by 27.73\%; kernel/runtime optimization and DVFS offset this increase, yielding a net reduction of 56.92\%.

\paragraph{Closed-loop effect of tree size.}
Figure~\ref{fig:hardware_ablation}c varies the leaf count $K$ of the same action-error-trained selector while keeping the four quantization configurations fixed.
Each setting uses the same 2,000 test episodes per model; Appendix~\ref{app:tree_success_ablation} gives the setup.
Both models perform best at $K=3$: 97.0\% for OpenVLA-OFT W4A8 and 98.4\% for $\pi_{0.5}$ W4A4.
These scores exceed $K=1$, the single-leaf selector with a training-selected fixed configuration, by 1.4 and 1.2 percentage points, and match the corresponding ActTune entries in Table~\ref{tab:quantization_performance}.
For both models, $K=4$ and $K=6$ score 0.6 and 0.4 points below $K=3$.
These closed-loop results support the three-leaf setting; increasing the number of leaves does not improve success in these comparisons.

\section{Related Work}
\label{sec:related_work}

\paragraph{Quantization and efficient VLA inference.}
Quantization and computation reuse reduce the cost of VLA inference.
GPTQ and AWQ calibrate low-bit weights~\citep{frantar2022gptq,lin2023awq}, while SmoothQuant and OmniQuant address weight--activation quantization through rescaling and reconstruction~\citep{xiao2022smoothquant,shao2024omniquant}.
For VLAs, QVLA, ActQuant, and task-evidence-aware mixed-precision quantization of VLA models (Mix-QVLA) use action sensitivity or task evidence to fit precision allocations~\citep{xu2026qvla,akbari2026actquant,ranjan2026mixqvla}; DyQ-VLA and MotionVQ adapt precision or quantized representations to motion during execution~\citep{zheng2026dyqvla,song2026vqvla}.
Other approaches improve decoding and action chunking~\citep{kim2025oft}, reuse visual tokens or intermediate features~\citep{xu2025vlacache,yang2025efficientvla}, or overlap inference with action execution~\citep{black2025rtc}.

\paragraph{Energy-aware inference and GPU operating-point control.}
Energy-aware deployment must also account for hardware operation and timing constraints~\citep{tang2019dvfs,stojkovic2024greener}.
Hardware-aware automated quantization (HAQ) uses hardware latency and energy feedback for bit allocation~\citep{wang2019haq}, and PiQi combines partial quantization, processor placement, and DVFS~\citep{aghapour2024piqi}.
DynamoLLM, a system for large language model (LLM) inference, adjusts serving resources and GPU frequency under latency objectives~\citep{stojkovic2025dynamollm}.
For robotic workloads, saliency-aware quantized imitation learning (SQIL) studies quantized imitation policies~\citep{park2025sqil}, while EcoVLA optimizes device-edge execution under latency constraints~\citep{zhou2026ecovla}.
We combine action-conditioned precision selection with measured GPU operating points and evaluate energy per successful episode, including failed attempts and idle intervals, while keeping the mean inference-latency increase within 10\% of the fixed reference operating point.

\section{Conclusion}
\label{sec:conclusion}

We presented ActTune, a framework for energy-efficient VLA inference through action-aware precision selection and GPU operating-point adaptation. Its decision tree learns splits and leaf configurations directly from configuration action errors, while hardware profiling identifies workload-specific energy--latency tradeoffs. Closed-loop LIBERO evaluation of this controller achieves 97.0\% mean success for OpenVLA-OFT W4A8 and 98.4\% for $\pi_{0.5}$ W4A4. With GPU operating-point adaptation, ActTune reduces energy per successful task by up to 76.8\% relative to the original BF16 implementations, while keeping the inference-latency increase below 10\% relative to the same quantized policies at the fixed reference operating point. Validation on physical robots remains future work.

\clearpage
\section*{AI Use Statement}
Generative AI tools assisted with implementing and revising analysis and visualization scripts, including aggregation of experimental summaries, calculation of speedups and relative energy savings, numerical consistency checks, and generation of figures and LaTeX tables. AI tools also assisted with literature searches, interpretation and presentation of experimental results, cross-reference checks, and language polishing. The authors take responsibility for the final code, experimental results, analyses, and manuscript.

\section*{Ethics Statement}
This work studies energy-efficient VLA inference using the LIBERO simulation benchmark. Reducing inference energy may lower the resource cost of robotic policies, but energy efficiency alone does not establish safe robot behavior. Deployment on physical robots requires task-specific safety assessment and validation beyond the simulation experiments reported here.

\section*{Reproducibility Statement}
The main text and appendices describe the model configurations, calibration procedure, evaluation protocol, and latency and energy measurements needed to reproduce the reported results.

\clearpage
\bibliography{iclr2027_conference}
\bibliographystyle{iclr2027_conference}
\clearpage
\appendix
\section{Appendix}
\label{app:experimental_details}

\subsection{Evaluation settings and result provenance}
\label{app:evaluation_settings}
All results in Table~\ref{tab:quantization_performance} are obtained from local evaluations, including the BF16 references and every quantization baseline. Success and energy evaluations use the Spatial, Object, Goal, and Long suites of LIBERO, with ten tasks and 50 initial states per task, giving 500 episodes per suite and 2,000 per model--precision setting.
We aggregate episodes by task and initial-state identity, counting each episode once even when a run is resumed.
The benchmark determines success, and we weight the four suites equally in the reported average.
All ActTune results use the action-error-trained decision tree in Sections~\ref{sec:action_class_recognition}--\ref{sec:shared_bank}, whose splits and leaf configurations are learned directly from configuration action errors. We use three leaves for the main results and vary this count only in the tree-size ablations. The selector is frozen before closed-loop evaluation and GPU profiling. Table~\ref{tab:quantization_performance} reports closed-loop success and energy for this method; Figure~\ref{fig:hardware_ablation}c examines closed-loop success as leaf count varies. Appendix~\ref{app:tree_leaf_ablation} reports the validation action loss and routing cost used to select tree size.
OpenVLA-OFT uses eight-action chunks.
The evaluated $\pi_{0.5}$ runtime executes ten actions per policy call and retains ten denoising steps.
These execution settings are distinct from the eight-action blocks used to define the diagnostic classes in Section~\ref{sec:motivation}.

For each model and suite, quantized methods start from the corresponding BF16 checkpoint and use the same 512-trajectory calibration selection described in Appendix~\ref{app:calibration}, while retaining their method-specific calibration objectives.
The cited papers define the compared methods; the table entries report our local measurements.
Each row reports one deployed system. Its success, inference latency, and energy come from the same 2,000 benchmark episodes with the same deployed checkpoint, calibrated configuration, and execution backend. Storage is measured from that deployed model, including all resident alternatives. Cross-method comparisons use the same task--initial-state pairs and hardware settings.
The $\Delta$ column is measured against the locally evaluated BF16 reference listed for the same model.
We compute means and $\Delta$ from unrounded suite averages, then round them to one decimal place using half-up rounding.

\paragraph{DyQ-VLA comparison scope.}
DyQ-VLA~\citep{zheng2026dyqvla} is closely related: it studies motion-dependent quantization sensitivity and uses kinematic signals for online precision switching.
At the time of this revision (September 2026), it is listed as an arXiv preprint, with no formal conference or journal publication reported. For this reason, we discuss it as related work but do not include it in the quantitative comparison in Table~\ref{tab:quantization_performance}.

\subsection{Quantization and execution scope}
\label{app:execution_scope}
Hardware measurements use NVIDIA RTX A6000 GPUs (Ampere, compute capability 8.6).
The selector uses 13 motion and gripper features plus current proprioception (Appendix~\ref{app:online_recognizer}). We set $\lambda_g=0.1$ in Equation~\ref{eq:action_calibration_loss}.

W4A8 and W4A4 denote the average precision budgets defined in Section~\ref{sec:shared_bank}.
Targeted layers can use two-, four-, or eight-bit weights, and W4A4 also allows two-, four-, or eight-bit activations.
The action head, proprioception projector, embeddings, normalization, and attention products retain their existing precision.
The two switchable matrices reuse resident alternatives, whose storage is included even when they are inactive.
The evaluated $\pi_{0.5}$ W4A4 implementation quantizes activations in groups of 64 channels.

Our W4A8/W4A4 inference measurements use the optimized integer backends in Section~\ref{sec:a6000_kernels}.
ActTune selects INT4$\times$INT4 or INT8$\times$INT8 kernels according to each layer's weight and activation precision. The W4A4 baselines use INT4$\times$INT4, and the W4A8 baselines use INT8$\times$INT8 with packed W4 weights. These deployed backends produce the reported success, latency, and energy measurements in the same closed-loop runs. Each energy-per-success denominator uses the successes recorded in those runs.
The local GPTQ and AWQ W4A4 results and the $\pi_{0.5}$ SmoothQuant W4A4 result use calibration that accounts for A4 activation quantization error.

\paragraph{GPU control and runtime checks.} The controller reads back the requested frequency and power cap after each update to verify that they were applied. Operating-point changes are separated by at least two inference requests. Runtime checks verify coverage of the targeted linear layers and compliance with the calibrated precision budgets.

\subsection{Retained weight/activation pairs}
\label{app:precision_inventory}
Table~\ref{tab:appendix_precision_inventory} lists the precision pairs retained in the evaluated banks.
For W4A8, the candidate set is $\{\mathrm{W2A8},\mathrm{W4A8},\mathrm{W8A8}\}$, and both models retain all three pairs.
For W4A4, the candidate space is $\{2,4,8\}\times\{2,4,8\}$, giving nine possible weight/activation pairs.
OpenVLA-OFT retains seven pairs, excluding W2A8 and W8A2; $\pi_{0.5}$ retains six, also excluding W8A8.
The retained pairs follow from the layer allocation described in Appendix~\ref{app:layer_precision_selection}.
Each W4A4 bank uses all three weight widths and all three activation widths, without requiring every pair.

\begin{table}[htbp]
\centering
\small
\setlength{\tabcolsep}{4pt}
\caption{Candidate and retained local weight/activation pairs in the frozen evaluated banks. Retained counts are distinct bit-width pairs, not action classes or whole-model configurations.}
\label{tab:appendix_precision_inventory}
\begin{tabular}{llccp{0.45\linewidth}}
\toprule
Model & Budget & Candidates & Retained & Retained pairs \\
\midrule
OpenVLA-OFT & W4A8 & 3 & 3 & W2A8, W4A8, W8A8 \\
OpenVLA-OFT & W4A4 & 9 & 7 & W2A2, W2A4, W4A2, W4A4, W4A8, W8A4, W8A8 \\
$\pi_{0.5}$ & W4A8 & 3 & 3 & W2A8, W4A8, W8A8 \\
$\pi_{0.5}$ & W4A4 & 9 & 6 & W2A2, W2A4, W4A2, W4A4, W4A8, W8A4 \\
\bottomrule
\end{tabular}
\end{table}

OpenVLA-OFT uses a separately calibrated bank for each LIBERO suite; the retained pair set is the same across all four suites, although the assigned layers differ.
Each $\pi_{0.5}$ precision setting uses one shared bank across the four suites.
Each of the four precision configurations in a bank contains the reported pair set.

The table counts distinct precision pairs across executed eligible linear matrices, excluding OpenVLA-OFT's unused vocabulary projection.
Each pair specifies one matrix's precision; a W8A8 matrix can therefore occur within a W4A4 average budget.
A whole-model configuration assigns a pair to every eligible matrix, and the precision selector chooses among four such configurations.
The accompanying precision inventory records the source checksums, per-configuration pair counts, weighted bit averages, and every layer's assignment in \texttt{layer\_assignments.csv}.

\subsection{Inference timing and model storage}
We record inference latency during the complete closed-loop benchmark runs used for the success and energy comparison. ActTune's selector uses the action history produced by those runs.
We use CUDA synchronization to time the complete prediction call, including input processing, configuration selection, policy inference, and action postprocessing.
Model loading, compilation, graph capture, remote procedure call (RPC) overhead, and simulation are excluded.
Measurements run on an RTX A6000 without competing computation. The no-DVFS comparison uses a fixed 1800\,MHz application-clock setting and 300\,W power cap; the DVFS comparison enables joint requested-frequency and power-cap adjustment, with the quantized policy fixed.

For suite $s$, let $\overline t_s^{\mathrm{BF16}}$ and $\overline t_s^q$ denote the mean prediction times of the original BF16 implementation and quantized configuration $q$.
The reported speedup is
\begin{equation}
S(q)=\frac{1}{4}\sum_s\frac{\overline t_s^{\mathrm{BF16}}}{\overline t_s^q}.
\label{eq:appendix_speedup}
\end{equation}
This average of per-suite ratios need not equal the ratio of two rounded overall mean latencies.
We use this aggregation formula for all methods. Initial task states match across methods, while subsequent trajectories may differ. The operating-point adaptation budget permits at most a 10\% increase in mean complete prediction-call latency relative to the same quantized policy at the fixed reference operating point (no DVFS). We check this budget using the inference timings recorded during the closed-loop benchmark.
The BF16 reference is the original implementation used in Table~\ref{tab:quantization_performance}; a separately optimized BF16 CUDA Graph implementation is outside this comparison.
The runtime optimizations preserve the frozen weights, precision assignments, selectors, and denoising steps.
Recorded actions match the reference exactly on the tested inputs, and model tensor bytes remain unchanged.
These action-equality checks cover the tested inputs; the full closed-loop runs provide the reported success and energy measurements.

Local model storage is the persistent compressed tensor payload in decimal GB ($10^9$ bytes).
It includes packed weight codes, quantization metadata, retained high-precision tensors, and every resident weight version, counting shared storage once.
For suite-specific checkpoints, our reported value is the maximum across suites.
Activation buffers, temporary workspaces, allocator caches, serialization overhead, and nonpersistent positional caches are excluded.
For SmoothQuant at W4A8 and W4A4, Table~\ref{tab:quantization_performance} reports the measured local W4 integer-backend storage: 6.32\,GB for OpenVLA-OFT and 3.67\,GB for $\pi_{0.5}$. The counts include packed weights, scales, offsets, input scales, persistent weight-sum buffers, and unquantized components such as the action head and proprioception projector. The shared packing layout gives the same static storage at both activation precisions.
For AWQ and GPTQ, we measure persistent storage from each deployed W4A8/W4A4 model using the same accounting: 5.91 and 6.01\,GB for OpenVLA-OFT, and 3.53 and 3.57\,GB for $\pi_{0.5}$, respectively. These counts include persistent VLA adapter tensors and unquantized components. The W4A8 and W4A4 deployments have the same reported storage at the table's precision; runtime activation buffers are excluded from this static measure.
Static model storage is distinct from sampled GPU memory; the table reports persistent tensor storage.

\subsection{Energy measurement}
Every energy entry in Table~\ref{tab:quantization_performance}, including BF16, W4A16 baselines, and ActTune with and without DVFS, uses the complete LIBERO benchmark: 500 episodes per suite and 2,000 per model--precision setting. Each method and operating-point setting uses the same task--initial-state identities; the subsequent closed-loop trajectories may differ.
The unparenthesized local W4A8/W4A4 energy entries use exclusive RTX A6000 runs at a fixed 1800\,MHz application-clock setting and a 300\,W power cap, with DVFS disabled. Parenthesized ActTune entries report the corresponding results with DVFS enabled.
We sample instantaneous GPU-board power through the NVIDIA Management Library (NVML) at approximately 20\,Hz and integrate each episode using the trapezoidal rule, from its first inference to its final environment step. The integral includes GPU rendering and intervals between policy calls; model loading, environment resets, explicit warmup, and central processing unit (CPU) energy are excluded.
Energy per success is the total energy of all 2,000 attempted episodes, including failures, divided by successes in those same runs. Runs with competing GPU processes are excluded and repeated.
This metric measures total GPU-board energy during simulated task execution, including inference, GPU rendering, and intervals between calls. It does not isolate inference energy or measure whole-robot energy.
The four-episode exploratory hardware sweep in Section~\ref{sec:motivation} uses a separate cohort and operating settings.

\subsection{GPU operating-point ablation cohort}
\label{app:hardware_ablation_cohort}
For the GPU operating-point ablation in Section~\ref{sec:ablation_study}, we use 400 evaluation episodes per setting: 100 from each of the Spatial, Object, Goal, and Long suites.
Within each suite, we randomly sample 100 episodes directly from its pool of 500 task--initial-state pairs (ten tasks with 50 initial states each), without imposing a per-task quota.
This subset is distinct from the full 2,000-episode success-rate evaluations in Table~\ref{tab:quantization_performance}.
The three-leaf precision selector is held fixed across the reference, best fixed, configuration-only, and region-and-configuration GPU operating-point policies.

\subsection{Closed-loop leaf-count ablation protocol}
\label{app:tree_success_ablation}
We test $K\in\{1,2,3,4,6\}$ using the calibration data and trajectory-level fitting/validation split in Appendix~\ref{app:calibration}.
For each model--precision pair, we learn tree splits and leaf configurations directly from the four configuration losses in Equation~\ref{eq:action_calibration_loss}, following Sections~\ref{sec:action_class_recognition}--\ref{sec:shared_bank} and Appendix~\ref{app:class_mapping_details}. Loss-reducing splits grow the initial tree with maximum depth five; bottom-up pruning produces a subtree with each requested leaf count $K$.
One tree shares its splits and configuration indices across the four suites. OFT resolves each index to the matching suite's bank; $\pi_{0.5}$ shares its bank across suites. Several leaves may select the same configuration. Model weights, the four-configuration banks, kernels, and fitting/validation split stay fixed across $K$.
At $K=1$, the single leaf selects the configuration index with the lowest weighted fitting loss. At $K>1$, current proprioception and previous-action features route each request to a leaf and its stored configuration. Missing history or invalid features trigger $q_{\mathrm{fallback}}=q_{11}$.
Here, $K$ is the number of leaves, as in Figure~\ref{fig:hardware_ablation}c. The PCA and K-means groups are used only for the Motivation diagnostics and do not supply labels for these trees.
Each tree is frozen before evaluation. The $K=3$ settings use the same selectors as the corresponding ActTune rows in Table~\ref{tab:quantization_performance}.
Each setting uses the same 500 task--initial-state pairs per suite for closed-loop tests, giving 10,000 episodes per model across all five settings. We report mean success over the four suites.

\section{Calibration following the QVLA protocol}
\label{app:calibration}

\subsection{Training data and held-out diagnostics}
Following the training-data source and trajectory-count convention in Appendix~F of QVLA~\citep{xu2026qvla}, we sample 512 distinct trajectories from the combined official LIBERO training tasks for calibration.
We use a frozen selection from the four suites with sampling seed 20260914.
The data snapshot is pinned to revision \texttt{f13aa24a3da8c43c7225569f28c562979fa0e35a} of the LIBERO dataset mirror.
Forty separate training trajectories are reserved for diagnostics; they and previously held-out diagnostic trajectories are excluded from calibration.
Evaluation rollouts are not used to fit quantization parameters or selectors.

Within each selected trajectory, we sample observations from frame zero at stride eight.
We exclude the final observation because it lacks a subsequent stored state for alignment checks.
This produces 10,996 calibration observations and 882 diagnostic observations (Table~\ref{tab:appendix_calibration}).
An observation at index $j$ is aligned with the next eight demonstration actions; near the trajectory end, the last available action is repeated to complete the stored block.
The manifest records trajectory and frame identities, observation hashes, and file checksums.
The shared-data loaders check disjoint trajectory identities, disjoint observation hashes, and the fixed frame ordering.

\begin{table}[htbp]
\centering
\small
\caption{Shared training calibration and held-out diagnostic data.}
\label{tab:appendix_calibration}
\begin{tabular}{lrrrr}
\toprule
Suite & \shortstack{Calibration\\trajectories} & \shortstack{Calibration\\frames} & \shortstack{Diagnostic\\trajectories} & \shortstack{Diagnostic\\frames} \\
\midrule
Spatial & 135 & 2,139 & 10 & 162 \\
Object & 121 & 2,341 & 10 & 185 \\
Goal & 138 & 2,305 & 10 & 174 \\
Long & 118 & 4,211 & 10 & 361 \\
\midrule
Total & 512 & 10,996 & 40 & 882 \\
\bottomrule
\end{tabular}
\end{table}

The stride, random seed, padding rule, and diagnostic split are explicit local choices.
QVLA does not provide the original trajectory identities or frame schedule, so we follow its data-source and 512-trajectory convention without claiming identical calibration samples.
Our shared manifest contains no additional instruction-only subset.
Following this calibration convention does not mean using QVLA's channel-wise quantizer. ActTune uses the layer-wise weight banks and precision budgets specified in Section~\ref{sec:shared_bank}.

\subsection{Calibration targets and separation from evaluation}
We use the selected observations to collect activation statistics, measure quantization error, and compare candidate configurations with BF16 reference actions.
ActTune calibrates candidate weight/activation settings and fits the configuration mapping using action agreement under the weight, activation, and resident-storage constraints of Section~\ref{sec:method}.
The local baseline adaptations retain their method-specific calibration objectives while using the shared training selection.
Layer coverage, quantizers, and execution backends can differ across methods despite the shared calibration set.
We freeze the settings before held-out diagnostics and benchmark evaluation, whose outcomes are excluded from fitting.

\subsection{Precision and storage budgets}
\label{app:precision_storage_budgets} Let \(\mathcal L\) index the targeted backbone linear matrices, with \(n_\ell\) parameters and \(v_\ell\) input activation elements per request. Each configuration \(q\in\mathcal Q\) assigns weight bits \(b_{\ell q}\in\{2,4,8\}\). W4A8 fixes activation bits \(a_{\ell q}=8\), whereas W4A4 allows \(a_{\ell q}\in\{2,4,8\}\). The budgets are
\begin{equation}
\begin{aligned}
\bar b(q)&=\frac{\sum_{\ell\in\mathcal L}n_\ell b_{\ell q}}{\sum_{\ell\in\mathcal L}n_\ell}\leq4,
&\bar a(q)&=\frac{\sum_{\ell\in\mathcal L}v_\ell a_{\ell q}}{\sum_{\ell\in\mathcal L}v_\ell}\leq4\quad\text{(W4A4)},\\
M_{\mathrm{bank}}&=\sum_{\ell\in\mathcal L}\sum_{(b,a)\in\mathcal B_\ell}\operatorname{bytes}(\ell,b,a)\leq M_{\max},
\end{aligned}
\label{eq:bank_constraints}
\end{equation}
where \(\mathcal B_\ell=\{(b_{\ell q},a_{\ell q}):q\in\mathcal Q\}\) contains the distinct resident versions of matrix \(\ell\). Storage includes packed weights, scales, and stored biases for every version, including inactive alternatives. The W4A4 activation bound also applies within each group of linear layers with the same invocation shape. Thus, average active precision and full-bank storage are separately constrained; W4A8 and W4A4 are budget labels, not uniform layer assignments. The action head, proprioception projector, embeddings, normalization, and attention products retain their existing precision; the unused vocabulary projection is excluded from \(\mathcal L\).

\subsection{Selecting precision for each layer}
\label{app:layer_precision_selection}
We select layer precision offline using reconstruction errors on training data, subject to bit and storage budgets.
Let $E_\ell(b,a)$ denote the action-Jacobian-weighted reconstruction error for quantizing eligible matrix $\ell$ to weight bits $b$ and activation bits $a$.
Calibration uses BF16 reference inputs and actions, activation-guided channel scaling, and row-wise weight clipping. For W4A4, we fit A4 scales and calibrate all nine weight/activation pairs from $\{2,4,8\}\times\{2,4,8\}$.
For $\pi_{0.5}$, the sensitivity calculation includes the ten-step denoising computation and the executed action coordinates.
This error is a proxy for ranking assignments; task success is evaluated separately in closed-loop rollouts.

Starting from uniform W4A8 or W4A4, the allocator assigns W2 to a capped set of matrices with the smallest increase in $E_\ell$ per parameter.
For W4A4, it also assigns A2 by ranking the error increase per input activation element within each invocation-shape group.
The predeclared W2 parameter fractions are $0.25\%$, $1\%$, and $2.5\%$.
For W4A4, these are paired with A2 input-element fractions of $2.5\%$, $5\%$, and $10\%$, respectively.
These fractions guide the initial candidate construction; subsequent storage repair can change the final assignments.

These reductions free bit budget for precision upgrades.
Writing $N=\sum_\ell n_\ell$ and $V=\sum_\ell v_\ell$, the greedy allocator ranks a feasible upgrade $(b,a)\rightarrow(b',a')$ by
\begin{equation}
\frac{E_\ell(b,a)-E_\ell(b',a')}
{(b'-b)n_\ell/N+(a'-a)v_\ell/V}.
\label{eq:appendix_precision_upgrade}
\end{equation}
The allocator accepts only upgrades that reduce the error proxy.
Every configuration must satisfy the parameter-weighted W4 bound; W4A4 must also satisfy the input-element-weighted A4 bound within each activation shape group.
For W4A8, activation precision stays at eight bits throughout allocation.
The greedy search enforces these bounds but does not establish a globally optimal layer assignment.

To limit the storage needed for configuration switching, we retain alternatives for two matrices.
We choose them from eligible lower-precision alternatives by the smallest reconstruction-error penalty, with a 2\,MiB limit per alternative backend at this construction stage.
For $\pi_{0.5}$, graph constraints keep vision-branch precision assignments fixed across configurations.
The two binary alternatives yield four whole-model configurations; all other matrices share one assignment across them.
The fourth configuration combines alternatives already stored for the three base configurations and requires no additional weight version.

The storage check includes all retained alternatives and their quantization metadata.
If the bank exceeds its byte limit, we reduce shared W8 matrices to W4, starting with the smallest error penalty per byte saved.

We select among feasible candidate banks by comparing their actions with BF16 on the calibration observations.
The score is mean continuous-action absolute error plus $0.1$ times gripper disagreement, averaged over the three base configurations.

The deployed W4A4 banks include model-specific refinements.
OpenVLA-OFT applies W4-to-W2 reductions ranked using training gradients to meet its compact-storage limit, then recomputes the bit counts and validates loaded storage.
The $\pi_{0.5}$ group-64 implementation reuses the calibrated candidate allocations and rescores them with grouped activation quantization before freezing the bank and selector.
At inference, the controller selects among the four frozen configurations, and the runtime applies their associated activation precisions.

Table~\ref{tab:appendix_switchable_layers} lists the two switchable matrices and their precision options in each evaluated bank; all other targeted matrices retain the same assignment across configurations.
\begin{table}[htbp]
\centering
\begingroup
\small
\setlength{\tabcolsep}{3pt}
\renewcommand{\arraystretch}{1.15}
\caption{Switchable layers in the evaluated banks. OFT denotes OpenVLA-OFT; V, L, and E denote its fused vision encoder, the $\pi_{0.5}$ language model, and the $\pi_{0.5}$ action expert, respectively. Block indices start at zero. Precision options correspond to the listed layers in order.}
\label{tab:appendix_switchable_layers}
\begin{tabular}{@{}llp{0.14\linewidth}p{0.34\linewidth}l@{}}
\toprule
Model & Budget & Suite & Switchable layers & Precision options \\
\midrule
OFT & W4A8 & Spatial, Object, Long
 & \shortstack[l]{\texttt{V.1.attn.qkv}\\\texttt{V.22.mlp.fc1}}
 & \shortstack[l]{W2A8 / W4A8\\W2A8 / W4A8} \\
OFT & W4A8 & Goal
 & \shortstack[l]{\texttt{V.19.attn.proj}\\\texttt{V.22.attn.proj}}
 & \shortstack[l]{W2A8 / W4A8\\W2A8 / W4A8} \\
\midrule
OFT & W4A4 & Spatial
 & \shortstack[l]{\texttt{V.1.attn.qkv}\\\texttt{V.23.mlp.fc1}}
 & \shortstack[l]{W2A2 / W4A2\\W2A2 / W4A2} \\
OFT & W4A4 & Object
 & \shortstack[l]{\texttt{V.17.attn.proj}\\\texttt{V.23.mlp.fc1}}
 & \shortstack[l]{W2A2 / W4A2\\W2A4 / W4A4} \\
OFT & W4A4 & Goal
 & \shortstack[l]{\texttt{V.17.attn.proj}\\\texttt{V.22.mlp.fc1}}
 & \shortstack[l]{W2A2 / W4A2\\W2A2 / W4A2} \\
OFT & W4A4 & Long
 & \shortstack[l]{\texttt{V.22.attn.proj}\\\texttt{V.23.mlp.fc1}}
 & \shortstack[l]{W2A4 / W4A4\\W2A2 / W4A2} \\
\midrule
$\pi_{0.5}$ & W4A8 & All four
 & \shortstack[l]{\texttt{L.16.self\_attn.o\_proj}\\\texttt{E.1.self\_attn.q\_proj}}
 & \shortstack[l]{W2A8 / W4A8\\W2A8 / W4A8} \\
$\pi_{0.5}$ & W4A4 & All four
 & \shortstack[l]{\texttt{L.15.self\_attn.q\_proj}\\\texttt{L.16.self\_attn.o\_proj}}
 & \shortstack[l]{W2A2 / W4A2\\W2A2 / W4A2} \\
\bottomrule
\end{tabular}
\endgroup
\end{table}

\subsection{Decision-tree inputs and diagnostic groups}
\label{app:online_recognizer}
The tree input is \(x_t=[\phi(A_{t-1},m_{t-1});s_t]\). Its 13 action features comprise translation and rotation magnitude, variation, curvature, and directional cancellation; combined-motion variation, curvature, and cancellation; and gripper transitions and margin. Current proprioception completes the input. Missing history or nonfinite features trigger \(q_{11}\).

The C0--C2 groups organize the Motivation diagnostics. They use standardized $8\times7$ action blocks, nine PCA components (95.97\% training variance), and K-means with $K=3$, seed 7, and 20 initializations. Training counts are 4,212/3,404/3,380. These diagnostic labels are separate from the decision regions learned from configuration errors.

\subsection{Adaptive tree fitting and trajectory-level validation}
\label{app:class_mapping_details}
We split calibration trajectories into fitting and validation subsets before calibrating the bank or tree. Fitting contexts provide the causal features above and four configuration losses from Equation~\ref{eq:action_calibration_loss}, measured with identical observations and policy random draws. Contexts without history use the fallback and are excluded from tree fitting. For context \(i\) in trajectory \(\tau\) and suite \(s\), we set \(w_i=1/(4M_s n_\tau)\), where \(M_s\) counts fitting trajectories with valid history in suite \(s\), and \(n_\tau\) counts valid fitting contexts in \(\tau\).

For fitting indices \(I\) at a node, its configuration and total loss are
\begin{equation}
q_I=\arg\min_{q\in\mathcal Q}\sum_{i\in I}w_i\ell_i(q),
\qquad R(I)=\sum_{i\in I}w_i\ell_i(q_I).
\label{eq:leaf_configuration}
\end{equation}
We choose the feature threshold maximizing \(R(I)-R(I_L)-R(I_R)\), requiring a positive reduction and at least 20 contexts per child. The initial tree has a depth cap of five. Let \(D_v(r)\) be the minimum fitting loss of a subtree rooted at node \(v\) with exactly \(r\) leaves. Bottom-up pruning sets \(D_v(1)=R(I_v)\) and, for \(r\geq2\), minimizes \(D_{v_L}(j)+D_{v_R}(r-j)\) over feasible child leaf counts \(j\). Infeasible counts have infinite cost. Backtracking from \(D_{\mathrm{root}}(K)\) yields the fitting-optimal subtree for each feasible leaf count \(K\), approximating Equation~\ref{eq:tree_calibration}; it is not a global search over all possible trees. The candidate set includes a single leaf and does not impose \(K=3\).

On held-out trajectories, we average loss within trajectories, then equally within and across suites, including the fallback at history-free requests. Validation measures the loss--routing-cost tradeoff across candidate sizes without changing the fitted splits or leaf choices. We use these curves to choose the experimental operating point; automatic one-standard-error selection is a separate exploratory comparison, not the rule used to obtain the final common three-leaf setting. Several leaves may share a configuration. The fixed tree and bank remain frozen for subsequent GPU profiling and evaluation.

\section{Supplementary results}
\label{app:supplementary_results}

\subsection{Leaf-count ablations across models}
\label{app:tree_leaf_ablation}
\begin{figure}[htbp]
\centering
\includegraphics[width=0.84\linewidth]{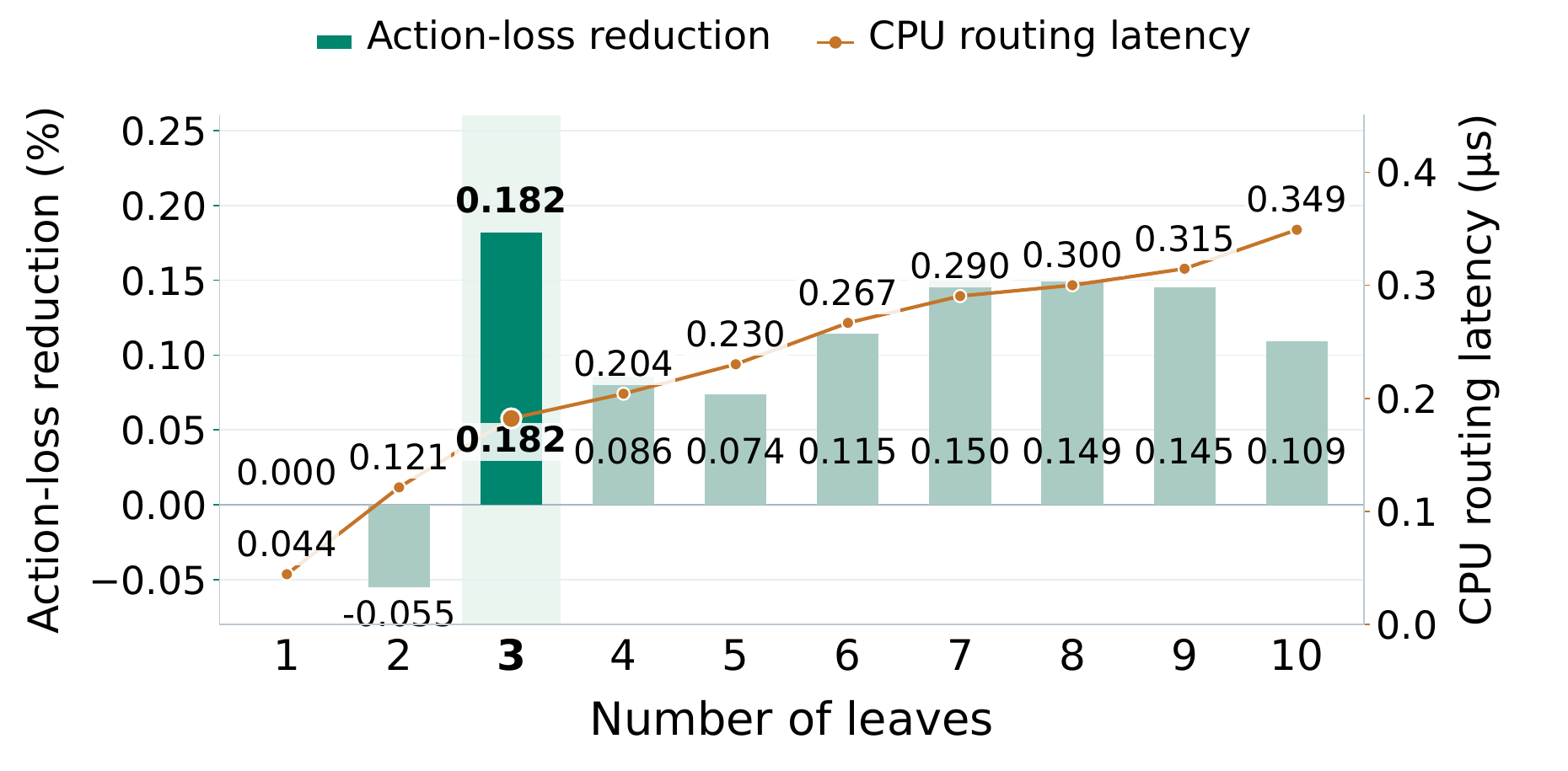}
\caption{OFT W4A8 leaf-count ablation for one tree shared across four LIBERO suites. Bars show validation action-loss reduction relative to one leaf; the line shows CPU routing latency. Three leaves give the lowest observed validation loss in this setting.}
\label{fig:appendix_oft_leaf_tradeoff}
\end{figure}

\paragraph{OFT W4A8.}
Figure~\ref{fig:appendix_oft_leaf_tradeoff} compares 1--10 leaves for the OFT W4A8 tree shared across four LIBERO suites.
Three leaves yield the lowest observed validation action loss, reducing it by 0.182\% relative to a single leaf, with a CPU tree-traversal latency of 0.182\,$\mu$s.
Every larger candidate has both higher loss and higher traversal latency.
Fewer leaves route faster but incur higher action loss. We therefore adopt three leaves as the common experimental setting based on this comparison and the tradeoffs across models.

\begin{figure}[p]
\centering
\includegraphics[width=0.84\linewidth]{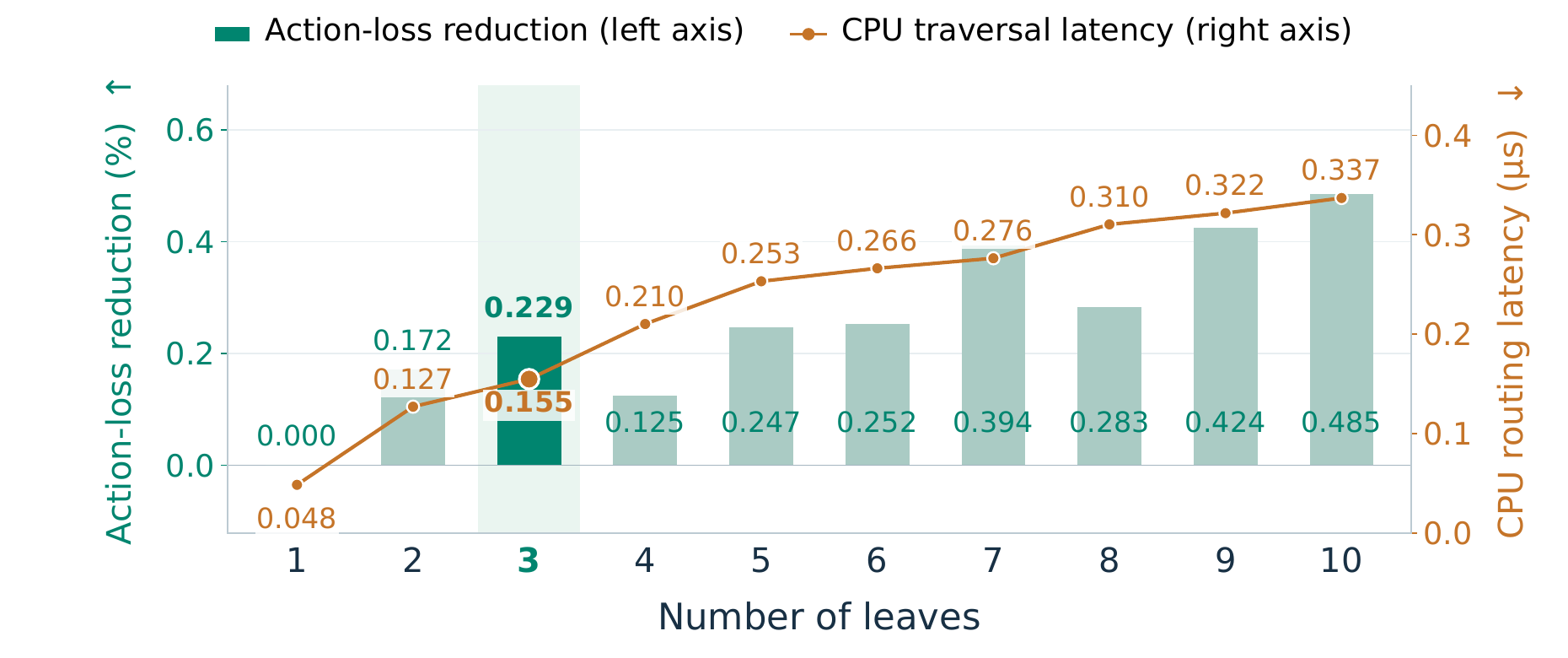}
\par\smallskip
{\small (a) OFT W4A4}
\par\medskip
\includegraphics[width=0.84\linewidth]{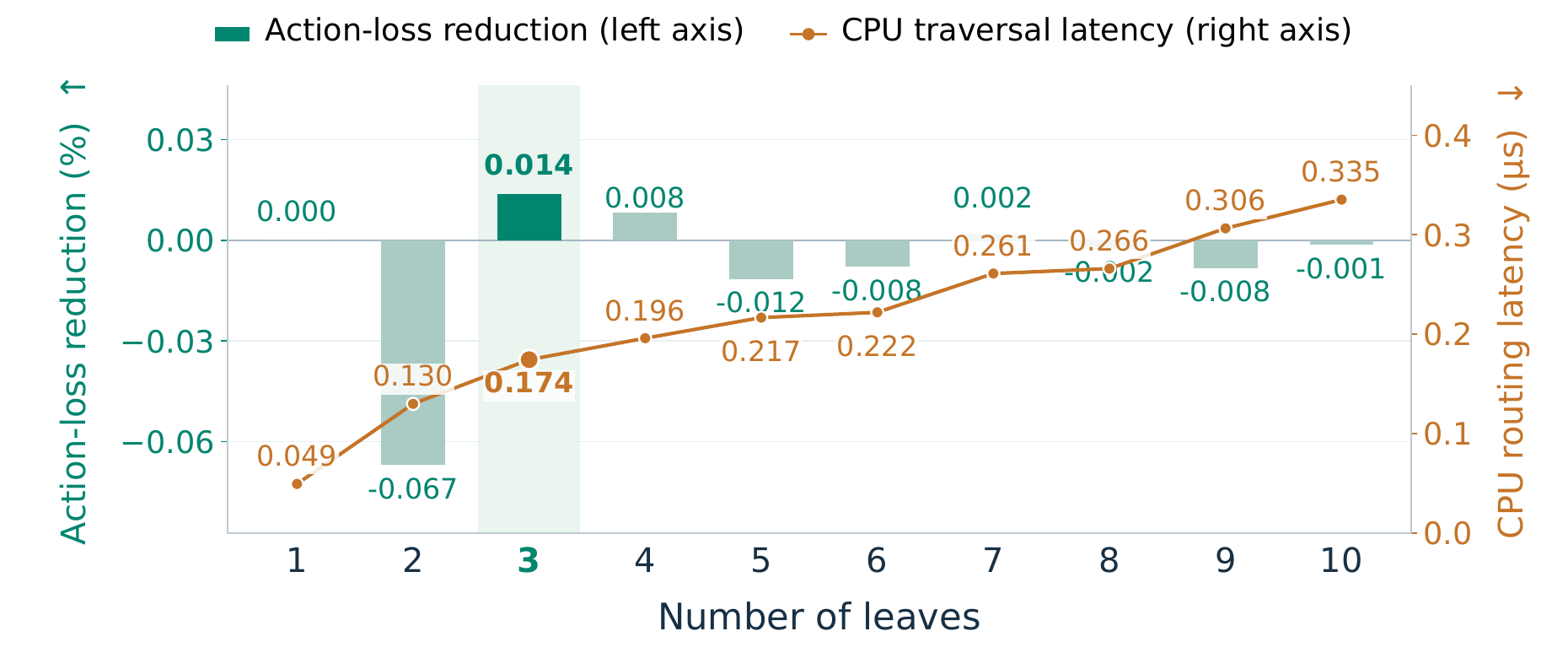}
\par\smallskip
{\small (b) $\pi_{0.5}$ W4A4}
\par\medskip
\includegraphics[width=0.84\linewidth]{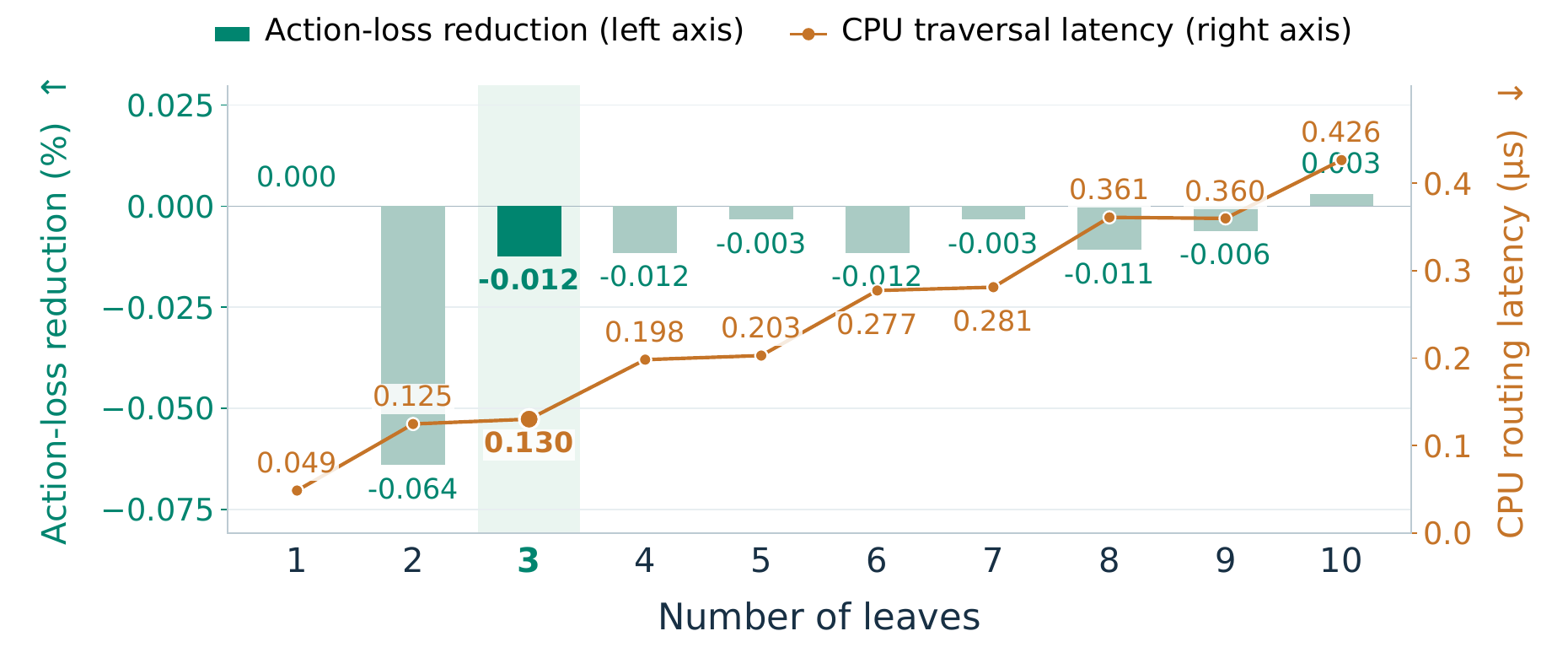}
\par\smallskip
{\small (c) $\pi_{0.5}$ W4A8}
\caption{Shared-tree leaf-count ablations for the remaining model--precision settings. Bars: validation action-loss reduction; lines: CPU routing latency. Three leaves are highlighted for comparison.}
\label{fig:appendix_tree_tradeoffs}
\end{figure}

\paragraph{Other model--precision settings.}
Figure~\ref{fig:appendix_tree_tradeoffs} extends Figure~\ref{fig:appendix_oft_leaf_tradeoff} to OFT W4A4 and $\pi_{0.5}$ W4A4/W4A8, using one tree shared across four LIBERO suites in each setting.
We evaluate every fixed leaf count from 1 to 10 by minimizing fitting loss over subtrees of the grown tree, including counts absent from the cost--complexity path, without refitting on validation data.
Action loss is averaged within each of 40 validation trajectories, then equally within and across suites; CPU traversal latency is measured over 21 rounds with the same protocol as Figure~\ref{fig:appendix_oft_leaf_tradeoff} and excludes model inference and action-feature extraction.
Within this range, the lowest observed validation loss occurs at 10 leaves for OFT W4A4, 3 for $\pi_{0.5}$ W4A4, and 10 for $\pi_{0.5}$ W4A8, with reductions of 0.485\%, 0.014\%, and 0.003\% relative to their single-leaf baselines.
These minima show that three leaves do not minimize loss in every setting. Offline loss alone does not establish a closed-loop benefit; Figure~\ref{fig:hardware_ablation}c reports the separate closed-loop leaf-count comparison for OFT W4A8 and $\pi_{0.5}$ W4A4.

\paragraph{Rationale for a common three-leaf setting.}
We adopt three leaves as a common experimental setting across model--precision pairs based on these validation curves. This choice retains context-dependent precision selection and requires at most two tree comparisons per decision.
For OFT W4A4, three leaves reduce action loss by 0.229\% relative to a single leaf; their loss is 0.257\% above the ten-leaf minimum, while traversal takes 0.155 rather than 0.337\,$\mu$s.
For $\pi_{0.5}$ W4A4, three leaves attain the lowest observed loss.
For $\pi_{0.5}$ W4A8, three leaves increase loss by 0.012\% relative to a single leaf and remain within 0.016\% of the ten-leaf minimum, with traversal taking 0.130 rather than 0.426\,$\mu$s.
The common three-leaf setting thus trades some models' minimum loss for lower routing cost; single-leaf routing remains faster. This comparison of fixed leaf counts is separate from automatic one-standard-error selection.

\subsection{Inference latency by suite}
Table~\ref{tab:appendix_latency} gives the per-suite times underlying the four ActTune speedups reported in Section~\ref{sec:quantization_performance} and the optimized latencies shown in Figure~\ref{fig:inference_runtime_comparison} (left). The speedups include runtime optimization and do not isolate quantization alone.

\begin{table}[htbp]
\centering
\small
\caption{Per-suite mean inference latency (ms) and speedup relative to the original BF16 runtime.}
\label{tab:appendix_latency}
\begin{tabular}{llrrrrrr}
\toprule
Model & Setting & Spatial & Object & Goal & Long & Mean & Speedup \\
\midrule
OpenVLA-OFT & BF16 & 149.21 & 150.69 & 148.71 & 151.52 & 150.03 & $1.00\times$ \\
OpenVLA-OFT & W4A8 & 99.76 & 99.65 & 99.47 & 101.31 & 100.05 & $1.50\times$ \\
OpenVLA-OFT & W4A4 & 98.50 & 98.08 & 98.16 & 97.87 & 98.15 & $1.53\times$ \\
\midrule
$\pi_{0.5}$ & BF16 & 239.00 & 243.90 & 243.76 & 247.59 & 243.56 & $1.00\times$ \\
$\pi_{0.5}$ & W4A8 & 119.72 & 120.38 & 120.22 & 121.23 & 120.39 & $2.02\times$ \\
$\pi_{0.5}$ & W4A4 & 133.97 & 134.15 & 135.47 & 135.69 & 134.82 & $1.81\times$ \\
\bottomrule
\end{tabular}
\end{table}

\subsection{Effect of inference optimization}
Table~\ref{tab:appendix_runtime_comparison} compares the earlier quantized runtime with the final implementation used in Table~\ref{tab:quantization_performance}. The earlier and final timings come from separate measurement campaigns; each reported speedup averages the four per-suite latency ratios.
BF16 denotes the original full-policy implementation used in Table~\ref{tab:quantization_performance}.
The archived records agree on input identities, predicted actions, selected profiles, and model tensor storage.

\begin{table}[htbp]
\centering
\small
\caption{Full-policy inference latency before and after runtime optimization (ms), with speedups relative to the earlier quantized runtime and original BF16 implementation.}
\label{tab:appendix_runtime_comparison}
\begin{tabular}{llrrrrr}
\toprule
Model & Setting & BF16 & Earlier & Optimized & \shortstack{Speedup\\vs. earlier} & \shortstack{Speedup\\vs. BF16} \\
\midrule
OpenVLA-OFT & W4A8 & 150.03 & 134.33 & 100.05 & $1.34\times$ & $1.50\times$ \\
OpenVLA-OFT & W4A4 & 150.03 & 100.32 & 98.15 & $1.02\times$ & $1.53\times$ \\
\midrule
$\pi_{0.5}$ & W4A8 & 243.56 & 373.39 & 120.39 & $3.10\times$ & $2.02\times$ \\
$\pi_{0.5}$ & W4A4 & 243.56 & 398.94 & 134.82 & $2.96\times$ & $1.81\times$ \\
\bottomrule
\end{tabular}
\end{table}

The comparison measures the combined effects of kernel and runtime optimization, including CUDA Graph replay, launch-geometry tuning, and model-specific execution changes.
It does not isolate the contribution of GEMM kernels.
The earlier OpenVLA-OFT W4A4 implementation already used language and vision graphs, leaving a smaller additional gain than the other configurations.
Matching actions on these recorded inputs is not a new closed-loop success-rate evaluation.

\subsection{Energy-saving contribution breakdown}
\label{app:energy_contributions}
Figure~\ref{fig:hardware_ablation}b is based on energy measurements from sequential component ablation experiments. For each model--precision setting, we measure GPU-board energy per successful episode for four variants: the BF16 reference, quantization only, quantization with kernel and runtime optimization, and the optimized quantized policy with GPU operating-point adaptation (DVFS). Table~\ref{tab:energy_contributions} reports the resulting contributions.
Each contribution is computed from the measured decrease in energy per successful episode between consecutive variants, divided by the corresponding measured BF16 reference energy per success and expressed in percentage points. Thus, the contributions add to the total saving up to rounding; they are not percentage reductions with a different denominator at each stage.
Kernel and runtime optimization are reported jointly. DVFS includes requested-frequency and power-cap adjustment, as defined in Section~\ref{sec:frequency_selection}.
Negative contributions denote increased energy. The $-27.73$-point quantization contribution for $\pi_{0.5}$ W4A4 means that this stage uses 127.73\% of BF16 energy per success. The subsequent 73.88-point kernel/runtime contribution offsets that increase, and the additional 10.77-point DVFS contribution yields a total energy saving of 56.92\%.
Energy depends on both average power and execution time. Lower power can therefore coexist with higher energy if execution takes sufficiently longer. For energy per success, failed attempts and the number of successes also affect the reported value.
\begin{table}[htbp]
\centering
\caption{Energy-saving contributions from sequential ablation measurements, with every component normalized by the corresponding measured BF16 energy per success. Component values are percentage points; total savings are percentages. The displayed components sum to the total saving in each row.}
\label{tab:energy_contributions}
\small
\begin{tabular}{llrrrr}
\toprule
Model & Precision & Quantization & \shortstack{Kernel\\+ runtime} & DVFS & Total (\%) \\
\midrule
OpenVLA-OFT & W4A8 & 49.82 & 9.71 & 10.68 & 70.21 \\
OpenVLA-OFT & W4A4 & 59.27 & 0.61 & 8.14 & 68.02 \\
$\pi_{0.5}$ & W4A8 & 28.09 & 42.82 & 5.92 & 76.83 \\
$\pi_{0.5}$ & W4A4 & $-27.73$ & 73.88 & 10.77 & 56.92 \\
\bottomrule
\end{tabular}
\end{table}

\subsection{Complete action-precision intervention grid}
Table~\ref{tab:appendix_action_grid} expands the selected comparisons in Figure~\ref{fig:motivation_action_combined}(b) to all nine weight/activation combinations.
For each of the ten tasks per suite, the experiment uses official initial state 1 and selects the first complete BF16 action block assigned to each class.
Every branch restores the same state, applies quantization for one eight-action block, and then returns to BF16 control until success or the original episode limit.
A BF16 sham branch verifies the reference continuation.
Each class--precision cell has ten paired branches per suite, except Goal C1 and C2, which have nine and eight because some reference trajectories contain no qualifying class window.
Missing windows are not imputed; success rates use the available paired branches within each suite, and the table weights suites equally.

Only the seven linear projections in each of the 32 language Transformer blocks are quantized in this intervention study.
Weights use the shared-training GPTQ group-128 calibration, while A2/A4/A8 use the same dynamic per-token affine quantization rule with 4/16/256 codes.
The vision modules and action head remain BF16.
These BF16 dense quantize--dequantize interventions diagnose sensitivity and are separate from ActTune's native integer-runtime evaluation.
Branches from one task share a reference episode and are not independent trials; the grid is a small diagnostic study, not the 500-episode-per-suite benchmark.

\begin{table}[htbp]
\centering
\small
\caption{Final success after one action-block precision intervention (\%). Entries average the four suites; the matched BF16 mean is 95.0\%.}
\label{tab:appendix_action_grid}
\begin{tabular}{lrrr}
\toprule
Intervention & C0 & C1 & C2 \\
\midrule
BF16 & 95.0 & 95.0 & 95.0 \\
\midrule
W2A2 & 62.5 & 35.0 & 91.9 \\
W2A4 & 67.5 & 42.5 & 95.0 \\
W2A8 & 97.5 & 97.5 & 94.4 \\
\midrule
W4A2 & 75.0 & 62.5 & 95.0 \\
W4A4 & 70.0 & 42.5 & 92.5 \\
W4A8 & 97.5 & 97.5 & 95.0 \\
\midrule
W8A2 & 75.0 & 52.5 & 95.0 \\
W8A4 & 67.5 & 45.0 & 92.5 \\
W8A8 & 97.5 & 97.5 & 95.0 \\
\bottomrule
\end{tabular}
\end{table}

With this small cohort and fixed quantizer, nonmonotonic entries do not establish a general advantage of lower precision or an ordering of C0--C2 by sensitivity.

\subsection{Action-grouping diagnostics}
On the 882 held-out blocks, the mean PCA-space silhouette is 0.1541, with 7.26\% negative values, indicating class overlap.
Across 20 refits that resample whole calibration trajectories within each suite, agreement with the frozen labels on these blocks has mean adjusted Rand index 0.8455 (range 0.7198--0.9829); all refits retain nine components.
These are descriptive separation and stability statistics, not classification accuracy or confidence intervals on task success.

\end{document}